\documentclass[10pt,twocolumn,letterpaper]{article}
\usepackage[pagenumbers]{cvpr}
\usepackage{cuted}   %
\usepackage{float}   %

\newcommand{\Tone}{$T_1w$}
\newcommand{\Ttwo}{$T_2w$}
\newcommand{\FLAIR}{$\mathrm{FLAIR}$}

\newcommand{\appref}[2]{\cref{#1}, \cref{#2}}   %

\renewcommand{\dbltopfraction}{0.8}

\AtEndPreamble{\crefname{appendix}{Appendix}{Appendices}\Crefname{appendix}{Appendix}{Appendices}}
\definecolor{cvprblue}{rgb}{0.21,0.49,0.74}
\usepackage[pagebackref,breaklinks,colorlinks,allcolors=cvprblue]{hyperref}
\def\confName{CVPR}
\def\confYear{2027}
\title{Pathology Randomization for Synthetic Priors and Zero-Shot Detection}
\usepackage{authblk}
\author[1]{Alexandra G. Roberts}
\author[1]{Tonmoy Hossain}
\author[1,2]{Mark Olchanyi}
\author[1,3]{Stefan Brandstätter}
\authorbreak
\author[1]{Jonathan Williams Ramirez}
\author[4]{Ana Lawry Aguila}
\author[1]{Matthew Rosen}
\authorbreak
\author[1]{Annabel Sorby-Adams}
\author[1]{W. Taylor Kimberly}
\author[1,2]{Juan Eugenio Iglesias}
\affil[1]{Harvard Medical School and Massachusetts General Hospital}
\affil[2]{Massachusetts Institute of Technology}
\affilbreak
\affil[3]{Medical University of Vienna}
\affil[4]{Microsoft}

\begin{document}
\maketitle
{\renewcommand{\thefootnote}{}\footnotetext{\raggedright Code: \href{https://github.com/agr78/Pathology-Randomizer}{\texttt{github.com/agr78/Pathology-Randomizer}}}}
\begin{strip}
\centering
\includegraphics[width=0.92\linewidth,height=0.4\textheight,keepaspectratio]{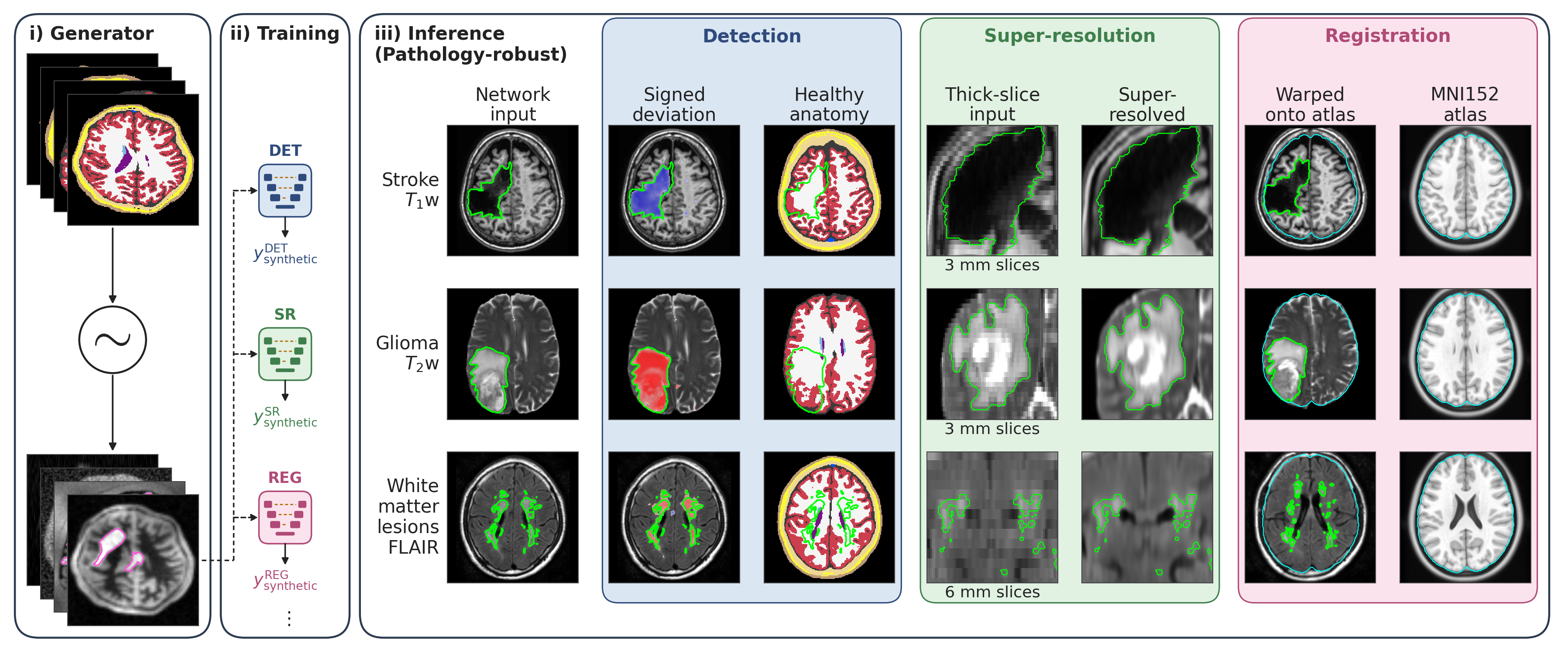}
\captionsetup{hypcap=false}
\captionof{figure}{\textbf{One generator for brain MRI with pathology.} (i) Pathology randomization draws lesions into healthy label maps (pink) and renders synthetic scans; (ii) these alone train one network per task; (iii) at inference, real scans with a stroke on \Tone{}, a glioma on \Ttwo{} and white matter lesions on \FLAIR{} (green: ground truth) are processed zero-shot: \emph{detection} (blue box) of the lesion as a signed deviation from the healthy anatomy predicted beneath it, \emph{super-resolution} (green box) from 3\,mm slices (6\,mm for the white matter lesions) to 1\,mm, and \emph{registration} (pink box) onto the \texttt{MNI152} atlas (cyan: atlas brain) with the lesion preserved. No real image, lesion mask or threshold is used in training.}
\captionsetup{hypcap=true}
\label{fig:schematic}%
\end{strip}

\begin{abstract}
Hospital brain imaging archives hold a diverse array of pathology at sample sizes beyond prospective studies, particularly for underrepresented populations and diseases. To unlock uncurated clinical data and scale analysis, methods are needed to detect abnormalities across contrasts and resolutions. Synthetic data has provided robust, contrast-agnostic models for a variety of human neuroimaging tasks in healthy anatomy, but pathology detection and other tasks in clinical settings still require real patient data, contrast-specific models, or disease-specific generators.
We propose \texttt{PathoDraw}, a generative model of brain MRI with pathology. A conditional distribution of lesions given the healthy anatomy is drawn into the label map before contrast-agnostic rendering, and a single network
estimates healthy anatomy with exact signed intensity deviation of any input scan.
The contrast-agnostic 3D UNet learns the healthy labels under the lesion and the signed deviation directly and identifies pathology via robust tissue statistics and learned features, despite no training on real pathology. We show zero-shot detection on public stroke, tumor, and white matter hyperintensity datasets as well as archived clinical data, spanning \Tone{}, \FLAIR{} and \Ttwo{} at various resolutions under a recent unsupervised anomaly detection benchmark protocol.
Our single model exceeds or performs comparably to contrast-specific anomaly detectors trained on real scans and keeps healthy anatomy intact.
The same generator trains networks for super-resolution and for registration in the presence of pathology. %
Our model, robust across acquisition and disease, %
enables the use of uncurated, archived data for clinical analysis. Our code is available at \href{https://github.com/agr78/Pathology-Randomizer}{\texttt{https://github.com/agr78/\allowbreak Pathology-Randomizer}}.
\end{abstract}

\section{Introduction}
\label{sec:intro}

Hospital archives hold orders of magnitude more brain MRI than any research cohort from a vast array of scanners, protocol, and diseases. Yet these data are uncurated, spanning a number of possible contrasts, resolutions and pathologies. Analyzing these data requires robustness to all three. Domain randomization over healthy label maps has addressed contrast and resolution variation with contrast- and resolution-agnostic segmentation, super-resolution, and registration of healthy MRI~\cite{synthseg,synthsr,synthmorph,anatomix,pentassuglia2026anatomy}. Pathology remains a critical gap, particularly when the disease population is not predefined. Synthetic pathology generators require real lesion masks or real pathological scans~\cite{strokesynth,una,synthpath,li2026brremark,wmhsynthseg,zhang2025mome} or disease-specific models~\cite{pidr,ren2026synhydro,cao2026haemorrhagic}, and state-of-the-art anomaly detectors are contrast specific~\cite{fae,disyre,disyrev2,andi,updstudy,uadbench,ma2026invdetect,baker2026patch2loc} or require real healthy or patient data~\cite{dgp,roy2026bayesian,deng2026fluxcut}.

We present a generative model of structural brain MRI with pathology. We model pathology as a conditional distribution of lesions given the healthy anatomy, drawn into the label map and rendered via domain randomization~\cite{tobin2017dr,tremblay2018dr}. With such ``pathology randomization'' we train a single contrast-agnostic network to estimate the healthy anatomy of any scan and where the scan deviates from it, zero-shot, without a real image, lesion mask or patient-derived threshold (\cref{fig:schematic}). We demonstrate robust inference across contrasts, resolutions, and disease status. %
Our contributions are:
\begin{itemize}[leftmargin=1.2em,itemsep=0pt,topsep=2pt,parsep=0pt]
  \item a disease-agnostic generative model of structural pathology for domain randomization: lesions drawn into healthy label maps, conditioned on the anatomy, with no real image or mask of any disease, and rendered contrast-agnostically (\cref{sec:prior}) and exact supervision  %
  (\cref{sec:network});
  \item zero-shot lesion detection on four datasets, three contrasts and 37 sites under the unsupervised anomaly detection (UAD) benchmark protocol
  (\cref{sec:results});
  \item generalization to zero-shot super-resolution and registration in the presence of pathology.
\end{itemize}

\subsection{Related work}
\label{sec:related}

The use of synthetic priors for healthy data is well-established. Segmentation~\cite{synthseg,pentassuglia2026anatomy}, super-resolution~\cite{synthsr} and registration~\cite{synthmorph} networks have been trained on renderings of healthy label maps, with extensions requiring real pathological data~\cite{synthseg,synthsr}. Synthetic data has also enabled feature learning~\cite{anatomix,diaz2026maskgen}, yet the label prior is healthy or requires real lesion segmentations~\cite{wmhsynthseg,billot2021ms}. \texttt{PathoDraw}, in contrast, keeps the prior synthetic and adds the pathology to it.

Segmentations from real pathology have informed synthetic data, as masks pasted into healthy label maps, labels under target-sequence priors, patient-fitted intensities, labeled susceptibility maps, or seeds for randomization~\cite{strokesynth,synthpath,li2026brremark,wmhsynthseg,roberts2025rim,pepsi,una}. \texttt{PathoDraw}, in contrast, needs no segmentation of real pathology: shape, placement and intensity are drawn from the prior.

\texttt{UNA} generates anomalies on label maps by advection--diffusion from real stroke annotations and noise seeds, and trains on real stroke scans among its synthetic ones. \texttt{PathoDraw} draws the lesion into the label map, conditioned on the anatomy and with mass effect, before rendering; supervises the exact signed deviation between the two renderings of one draw and the healthy labels beneath; and admits no real image, mask or threshold. Additionally, \texttt{PathoDraw} is evaluated on stroke, tumors, and white matter hyperintensities, enabling super-resolution and registration from the same generator.

Disease-specific generators model one process, such as corpus-callosum malformations (\texttt{PIDR}~\cite{pidr}), biomechanical hydrocephalus (\texttt{SynHydro}~\cite{ren2026synhydro}) or procedural hemorrhage~\cite{cao2026haemorrhagic}, needing no real pathology but limited to the disease of interest. \texttt{PathoDraw} is tied to no disease process; one prior serves stroke, glioma and white matter hyperintensities.

Like our method, anomaly detection does not require real pathology data. Feature-space reconstruction (\texttt{FAE}~\cite{fae}), cold-diffusion restoration of painted anomalies (\texttt{DISYRE}~\cite{disyre,disyrev2}), diffusion-latent scoring~\cite{ma2026invdetect}, patch-location prediction~\cite{baker2026patch2loc}, disentangled anatomy~\cite{yang2026disentangled} and training-free foundation-model scoring~\cite{legia2026trainingfree} generalize across pathologies, training on real healthy images and primarily using one model per contrast~\cite{updstudy,uadbench,behrendt2026uadreview}. That painted-anomaly line, from patch interpolation~\cite{fpi,pii,nsa} to pasted or textured patches~\cite{cutpaste,draem,baugh2023} and \texttt{DISYRE}, corrupts real healthy images in pixel space and binds each model to one contrast; scoring voxels as outliers of a per-class tissue model goes back to~\cite{vanleemput2001}. Diffusion priors over real healthy images solve detection as a per-scan inverse problem (\texttt{dmprior}~\cite{dgp}) or, in 2D with one prior per contrast, with a jointly sampled latent mask~\cite{roy2026bayesian}, healthy counterfactuals produce anomaly maps when subtracted.

None of these methods accept arbitrary clinical scans. The anomaly detectors generally require one model per contrast.
Synthetic-pathology generators either require real pathology data or model one disease process. To our knowledge, no prior generator expresses entirely synthetic, disease-agnostic structural pathology, formulated as a domain-randomization problem.

\section{Method}
\label{sec:method}

\begin{figure*}[t]
  \centering
  \includegraphics[width=\linewidth,height=0.6\textheight,keepaspectratio]{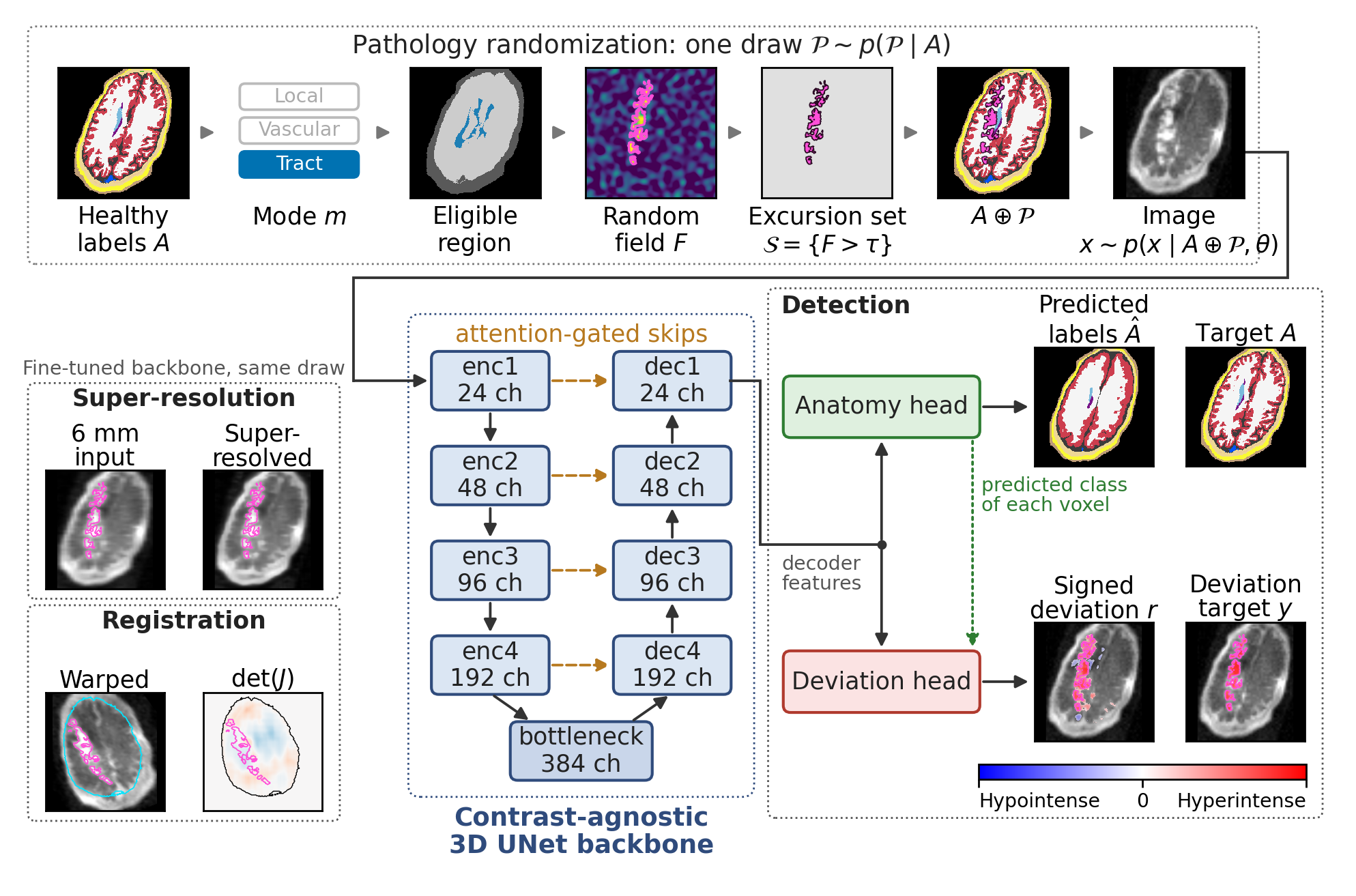}
  \caption{\textbf{Pathology randomization and the network}, one draw of the generative model. The top row shows the healthy labels, the drawn mode and its eligible region, the random field whose excursion set is the lesion, the injected label map (pink) and the rendering (\appref{tab:generator}{sec:suppl_generator}). Below, the rendering enters the contrast-agnostic 3D UNet backbone, whose anatomy head predicts the healthy labels and deviation head the signed deviation. The same draw passes through the task networks, super-resolved from 6\,mm slices and registered to another draw's healthy rendering, with the warped lesion (pink), target brain (cyan) and Jacobian determinant.}
  \label{fig:flow}
\end{figure*}

\begin{figure*}[t]
  \centering
  \IfFileExists{figs/teaser_bestcase.png}{\includegraphics[width=\linewidth,height=0.42\textheight,keepaspectratio]{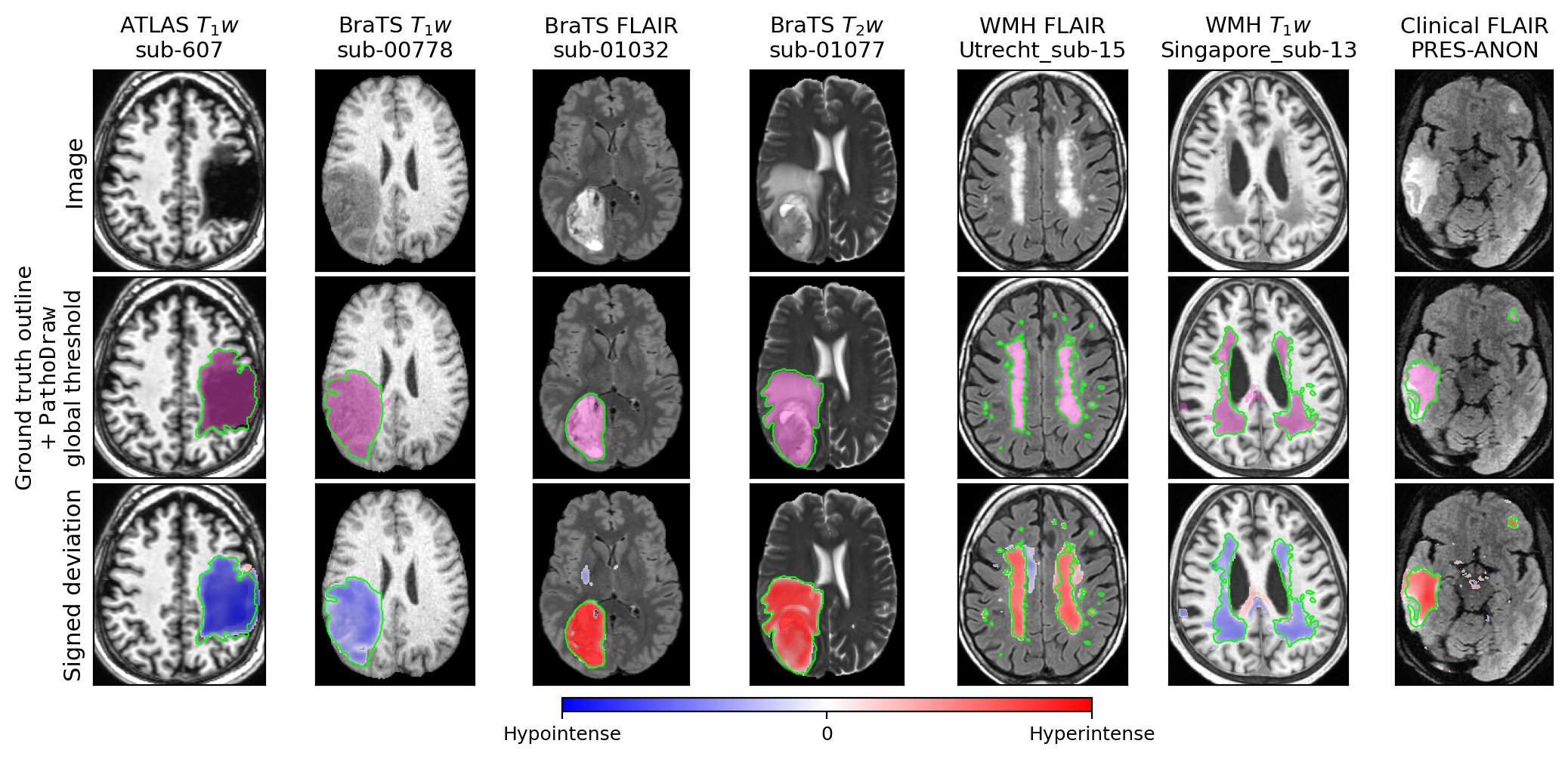}}{\fbox{teaser pending}}
  \caption{\textbf{Exemplar cases}. One model on test scans from four diseases, three contrasts and 37 sites; one column per dataset (ATLAS \Tone{}, BraTS \Tone{}/\FLAIR{}/\Ttwo{}, WMH \FLAIR{}/\Tone{}, clinical posterior reversible encephalopathy syndrome (PRES) \FLAIR{}),
  with the input image, ground truth (green) and our segmentation (pink), and the signed deviation map, at protocols from 1\,mm isotropic to 6\,mm slices.}
  \label{fig:teaser}
\end{figure*}

Our framework consists of a generative model of a structural brain MRI scan with pathology. We begin with a model of healthy anatomy, a conditional distribution of pathology given that anatomy, and a renderer that turns the combined label map into an image under a randomized acquisition. We define the model (\cref{sec:prior,sec:anatomy,sec:prior-labels,sec:intensity}), the anatomy factorization the network estimates (\cref{sec:targets}), and a network fit to recover it from samples of the model, since the posterior itself is intractable (\cref{sec:network}).

\subsection{Generative model}
\label{sec:prior}

A scan $x$ is generated from a healthy label map $A$, a pathology $\mathcal{P}$ and an acquisition $\theta$, with joint distribution
  \begin{equation}
    p(x,A,\mathcal{P},\theta)=p(x\mid A,\mathcal{P},\theta)\,p(\mathcal{P}\mid A)\,p(A)\,p(\theta).
    \label{eq:gen}
  \end{equation}
The factorization follows from two independence assumptions, $A\perp\theta$ and $\mathcal{P}\perp\theta\mid A$ (\cref{sec:suppl_joint}). Here $p(A)$ is the empirical distribution of healthy segmentations under spatial augmentation (\cref{sec:anatomy}). The distribution $p(\mathcal{P}\mid A)$ places a lesion, with its own labels and a mass-effect deformation, on that anatomy (\cref{sec:prior-labels}). The prior $p(\theta)$ is uniform over wide ranges of tissue contrast, resolution, bias field and noise (domain randomization~\cite{synthseg}). Simulated $p(x\mid A,\mathcal{P},\theta)$ renders the label map with the pathology injected (\cref{sec:intensity}). Rendering the same $A$ under the same $\theta$ with and without $\mathcal{P}$ gives a pathological image and its control, the supervision of \cref{sec:targets}. Every probability and range below is fixed a priori (\appref{tab:generator}{sec:suppl_generator}).

\subsection{Model of healthy anatomy}
\label{sec:anatomy}

We assume a dataset of $N$ 3D segmentations $\{A_i\}$ of healthy subjects into $K=56$ labels (\cref{sec:experiments}). A draw is one of them under a random spatial transform,
  \begin{equation}
    i\sim\mathcal{U}\{1,\dots,N\},\qquad \varphi\sim p(\varphi),\qquad A=A_i\circ\varphi^{-1},
    \label{eq:anat}
  \end{equation}
where $\varphi$ composes a random affine transform, a smooth nonlinear field and a left--right flip with a $160^3$ crop at a random center, as in~\cite{synthseg} prior to intensity draws.

\subsection{Pathology randomization}
\label{sec:prior-labels}

The conditional distribution $p(\mathcal{P}\mid A)$ is a mixture of three disease modes chosen to span the geometry of structural brain pathology rather than any one disease~\cite{strive}. We introduce tract-aligned lesions in the periventricular white matter (small-vessel disease), vascular-aligned lesions in the deep gray nuclei or along a cortical territory (infarcts), and local lesions in any brain label (tumors, focal lesions), amenable to mode additions. A pathology is constructed by a sequence of draws from random variables. We term this pathology randomization method \texttt{PathoDraw}. A primary mode $m\sim\operatorname{Cat}(\pi_{\mathrm{local}},\pi_{\mathrm{vasc}},\pi_{\mathrm{tract}})$ is drawn; each remaining mode then adds a second pathology, independently, with probability $0.15$. %
  \begin{equation}
    \mathcal{S}_m=\{v\in\Omega_m(A):W_m(v)F_m(v)>\tau_m\},\; \mathcal{S}=\textstyle\bigcup_m\mathcal{S}_m,
    \label{eq:support}
  \end{equation}
with $\tau_m$ the quantile that keeps a random fraction of the envelope. Eroded into a core and a shell, each a new label $\ell_{\mathcal{P}}(v)$, the pathology is injected by ``overwriting,''
  \begin{equation}
    (A\oplus\mathcal{P})(v)=\begin{cases}\ell_{\mathcal{P}}(v)& v\in\mathcal{S},\\ A(v)& v\notin\mathcal{S},\end{cases}
    \label{eq:inject}
  \end{equation}
a hard replacement of labels rather than a blend of images, since every intensity is drawn afterwards (\cref{sec:intensity}). Mass effect~\cite{prastawa2009} displaces the surrounding anatomy along the smoothed, normalized gradient of the lesion field, $\psi=\mathbb{I}+\alpha\,\nabla(G_\sigma\ast F_m)/\|\nabla(G_\sigma\ast F_m)\|$, with a mode-specific probability and a magnitude $\alpha$ clamped to a fold-free range. Thus $\mathcal{P}=(\mathcal{S},\ell_{\mathcal{P}},\psi)$, with support $L=\mathcal{S}$ the lesion label of \cref{sec:targets}. The control label map is the anatomy $A$. \Cref{fig:flow} shows one draw with details in \cref{sec:suppl_prior}.

\subsection{Rendering}
\label{sec:intensity}

The acquisition $\theta=(\mu,\sigma,\beta,r,\eta)$ holds one mean and one standard deviation per label, a bias field, a resolution and a noise level, each drawn uniformly over a wide range~\cite{synthseg}. With $L_{\mathcal{P}}=(A\oplus\mathcal{P})\circ\psi^{-1}$ the pathological label map, the image is rendered one stage at a time,
  \begin{align}
    \mu_l&\sim\mathcal{U}[a_{\mu,l},\, b_{\mu,l}], &&\sigma_l\sim\mathcal{U}[a_{\sigma,l},\,b_{\sigma,l}],\notag\\
    G(v)&=\big(\mu_{L_{\mathcal{P}}(v)}+\sigma_{L_{\mathcal{P}}(v)}\, \xi(v)\big)\,e^{\beta(v)}, &&\xi(v)\sim\mathcal{N}(0,1),\notag\\
    x&=\mathcal{D}_r(G)+\varepsilon, &&\varepsilon\sim\mathcal{N}(0,\eta^2),
    \label{eq:render}   %
  \end{align}
so that $x\mid L_{\mathcal{P}},\theta$ is a Gaussian mixture conditioned on the labels alone, with the bounds $a_{\mu,l}=25$, $b_{\mu,l}=225$, $a_{\sigma,l}=5$, $b_{\sigma,l}=25$ of \texttt{SynthSeg}~\cite{synthseg,synthsegcode} for every label $l$. $G$ is the 1\,mm rendering with partial-volume ramps at label boundaries and a smooth multiplicative bias field $e^{\beta}$. The operator $\mathcal{D}_r$ blurs $G$ with the slice profile of resolution $r$ (isotropic, or thick slices along one axis), subsamples it to $r$ and resamples back to 1\,mm, and $\varepsilon$ is acquisition noise. The control image $x^{0}$ ($G^{0}$ at 1\,mm) is rendered from $A$ with the same $\theta$, $\xi$ and $\varepsilon$, so the two renderings differ only where the labels differ. Lesion labels obey the following conventions. (i) They are rendered by composition between the host tissue they overwrote and a modulating class, CSF or gray matter, $\mu_{\ell}=(1-f)\,\mu_{\mathrm{host}}+f\,\mu_{\mathrm{mod}}$ with $f$ uniform in a per-mode range, so that lesion polarity follows the drawn contrast, dark where white matter renders bright (\Tone{}-like draws) and bright otherwise (\FLAIR{}-like draws). (ii) Every lesion satisfies the Rose criterion~\cite{rose1948}, in that its mean rendered change against the control over $\mathcal{S}$ is held between $k\,\sigma_{\partial\mathcal{S}}$ and $2k\,\sigma_{\partial\mathcal{S}}$, $k\sim\mathcal{U}[1,2.5]$, with $\sigma_{\partial\mathcal{S}}$ the control's standard deviation in a band around the lesion, so that no invisible lesion becomes a training target (\cref{sec:suppl_rose}). (iii) Dark-CSF contrast~\cite{hajnal1992,rooney2007,wansapura1999,mugler1990} on half of the draws (\cref{sec:suppl_ablation}).

\subsection{Anatomy factorization}
\label{sec:targets}

The network factorizes a scan into the healthy anatomy as if the pathology were absent and a deviation map of where and how the scan departs from it. Lesion detection is posed as a global thresholding rule on that map. The anatomy target is $A$, the healthy labels under the lesion. The deviation target is built from the two renderings of one draw, $\Delta I=G-G^{0}$, in the scan's own intensity frame $s=\wp_{99}-\wp_{1}$, the percentile range synthetic-data methods rescale their inputs by~\cite{synthseg,synthsr,synthmorph}. On the lesion support $L$, $y_v=\operatorname{sgn}(\Delta I_v)\,\max\!\big(|\tanh(\Delta I_v/s)|,\,C\big)$, and $y_v=0$ outside it, with signal floor $C=0.18$ from the Rose criterion (\cref{sec:suppl_rose}). The magnitude $|y|$ is a soft lesion mask that never vanishes inside a lesion drawn as visible, its sign the lesion's polarity, and both are zero on the control image.

\subsection{Inference with a neural network}
\label{sec:network}

The posterior of anatomy and pathology under \cref{eq:gen},
$p(A,\mathcal{P}\mid x)\propto p(A)\,p(\mathcal{P}\mid A)\int p(x\mid A,\mathcal{P},\theta)\,p(\theta)\,d\theta$,
is intractable, since the acquisition $\theta$ must be marginalized. Sampling the forward model, however, is tractable, and every draw comes with respective latent variables: the healthy labels and deviation of \cref{sec:targets}, the 1\,mm rendering, and the deformation between two draws, which are the targets of detection, super-resolution and registration. Tasks $\tau$ like detection, super-resolution, and registration, whose targets $t_\tau=t_\tau(A,\mathcal{P},\theta)$ are such draw-derived quantities, each with a loss $\mathcal{L}_\tau$, can therefore be learned by amortized inference, a network $g_\phi$ fit on draws,
  \begin{equation}
    \phi^\star_\tau=\operatorname*{arg\,min}_\phi\;\mathbb{E}_{A,\mathcal{P},\theta,\xi,\varepsilon}\big[\mathcal{L}_\tau\big(g_\phi(x),\,t_\tau(A,\mathcal{P},\theta)\big)\big],
    \label{eq:amortized}
  \end{equation}
which at the optimum is the Bayes estimator of $t_\tau$ under $\mathcal{L}_\tau$~\cite{cranmer2020}. The generative model is the same for every task, only the pair $(t_\tau,\mathcal{L}_\tau)$ changes, and no real image is used. The targets are $t_{\mathrm{DET}}=(A,y)$ of \cref{sec:targets}, $t_{\mathrm{SR}}$ the 1\,mm rendering, and $t_{\mathrm{REG}}$ the label overlap between two draws, with the losses $\mathcal{L}_{\mathrm{DET}}$, $\mathcal{L}_{\mathrm{SR}}$ and $\mathcal{L}_{\mathrm{REG}}$ given in \cref{sec:tasks}.

\noindent \textbf{Network.} $g_\phi$ is a 3D UNet (\cref{fig:flow}) after~\cite{synthseg} with identity-initialized attention gates on the skips~\cite{attnunet,olchanyi2026brainstem}. The same backbone serves every task behind a task-specific first convolution and head. For detection the head is dual. The anatomy head ($1\times1\to56$ classes, softmax) predicts the healthy label map, and the deviation head is a two-layer $1\times1$ MLP on the 24 decoder features plus $\tilde x$, the input rescaled by the median and median absolute deviation (MAD) of its anatomy-head class (\cref{sec:suppl_statistic}). The two task gradients on the backbone are balanced per step~\cite{imtl}. \label{sec:readout}The detection readout is the deviation head's magnitude $|r|$. The super-resolution head predicts a residual on the input and a per-voxel amplitude in $(0,1)$ that scales it, and the registration head a diffeomorphic deformation field, each with its final convolution zeroed to start at the identity.

\noindent \textbf{Controls.} Every task has a control whose backbone was trained without pathology in the draws, so that pathology enters only through the training data (\cref{sec:suppl_downstream}).

\subsection{Task losses}
\label{sec:tasks}

\noindent \textbf{Detection.}\label{sec:loss} Over the batch holding the control and the pathological image of one draw, $\mathcal{L}_{\mathrm{DET}}=\mathcal{L}_{\mathrm{anat}}+\mathcal{L}_{\mathrm{dev}}$, with $\mathcal{L}_{\mathrm{anat}}=\mathcal{L}^{\mathrm{ROI}}_{\mathrm{CE}}+\mathcal{L}_{\mathrm{Dice}}$~\cite{nnunet,synthseg,wmhsynthseg} on the anatomy target $A$ and
$\mathcal{L}_{\mathrm{dev}}$, an $L_1$ loss on the deviation head: on the pathological rendering, $|\hat r_v-y_v|$ weighted to balance the lesion against the background and to stress the border band where a detector oversegments, and on the control, $|\hat r^{0}_v|$, where the deviation must vanish. The cross-entropy is weighted likewise inside the ROI, the healthy structures the lesion overlaps; the weights are fixed a priori after~\cite{unet,dbce} and written out in \cref{sec:suppl_loss}.

\noindent \textbf{Super-resolution.}\label{sec:downstream} For $\mathcal{L}_{\mathrm{SR}}$ the network predicts a residual $f(\bar x)$ on the generator's own degraded image using operator $\mathcal{D}_r$, $x=\mathcal{D}_r(x^{\mathrm{hr}})+\varepsilon$ with $x^{\mathrm{hr}}=G^{0}$ the 1\,mm rendering before the acquisition stage, and is fit with an $L_1$ loss, $\mathcal{L}_{\mathrm{SR}}=\tfrac{1}{|B|}\sum_{v\in B}\big|\bar x_v+f(\bar x)_v-\bar x^{\mathrm{hr}}_v\big|$ over the brain mask $B$ of $A$, where $\bar x=(x-\wp_{1}(x))/(\wp_{99}(x)-\wp_{1}(x))$ and $\bar x^{\mathrm{hr}}$ is scaled by the same percentiles of the input, as in \cref{sec:targets}. At test time every scan is resampled to 1\,mm before the network, so any voxel size is accepted.

\noindent \textbf{Registration.} For $\mathcal{L}_{\mathrm{REG}}$ the network takes the moving and fixed image as two channels and predicts a stationary velocity field $\nu=g_\phi(x_a,x_b)$, integrated by scaling and squaring 
into a diffeomorphic field $\varphi=\exp(\nu)=\mathbb{I}+u$. It is fit with \texttt{SynthMorph}'s loss~\cite{synthmorph,synthmorph_joint}, $\mathcal{L}_{\mathrm{REG}}=1-\mathrm{Dice}(A_a\circ\varphi,\,A_b)+\lambda\,\tfrac{1}{3|V|}\sum_{v\in V}\|\nabla u_v\|_F^{2}$ with $\lambda=1$, the soft Dice taken over the brain labels present in the pair and $\nabla u$ the forward-difference gradient of the displacement in voxel units over the volume $V$.
The pair is two independent draws under unique spatial augmentation (\cref{sec:suppl_registration}). The moving image is the pathological rendering with probability $\tfrac12$ and the fixed image is the control rendering. The control trains likewise on healthy renderings only.

\section{Experimental setup}
\label{sec:experiments}

\subsection{Anatomy source}

We draw healthy anatomy uniformly from 1{,}033 label maps of young adults of the Human Connectome Project (HCP-YA~\cite{hcp}) and the Chinese HCP~\cite{chinesehcp}, segmented with \texttt{FreeSurfer}~\cite{freesurfer} under manual quality control, augmented with the claustrum and limbic structures by the label pipeline of~\cite{reffree} and with extracerebral labels from \texttt{SAMSEG}~\cite{samseg} and \texttt{CHARM}~\cite{charm}, merged to the 56-label scheme of the anatomy head. No image of these subjects enters training. %

\subsection{Test data}

We infer on the following datasets: \textbf{ATLAS}~\cite{atlas} (acute to chronic stroke, \Tone{}), the benchmark's unseen-scanner subset ATLAS-O ($n{=}210$, nine scanner cohorts). \textbf{BraTS}~\cite{brats2021} ($n{=}1{,}249$ from 24 sites~\cite{brats2021tcia}, \Tone{}, \FLAIR{}, \Ttwo{} co-registered) is the 2023 adult glioma pool, scored as whole tumor on every modality~\cite{uadbench}. \textbf{WMH}~\cite{kuijf2019} ($n{=}60$) has three sites, with masks delineated on \FLAIR{} and reported on \Tone{} by the benchmark~\cite{uadbench}. \textbf{PRES} (posterior reversible encephalopathy syndrome, $n{=}43$) is a hospital-archive cohort, clinical \FLAIR{} at acquired ${\approx}0.5{\times}0.5{\times}6$\,mm, with manual masks. These scans were acquired during routine clinical care and analyzed retrospectively as secondary data under Institutional Review Board approval at our institution, with informed consent obtained. \textbf{ADNI3}~\cite{adni3} ($n{=}331$ per contrast, \Tone{} and \FLAIR{}; 132 cognitively normal, 199 impaired) is a neurodegenerative cohort without focal lesions.

\subsection{Evaluation}

\noindent \textbf{Performance metrics}. The published benchmark's primary metric~\cite{uadbench} is the ceiling Dice $\lceil$Dice$\rceil$ (\cref{tab:main}), a global threshold per dataset chosen on the scored cases (an oracle, hence an upper bound), 4\,mm median filter, per-case Dice averaged over every case inside a \texttt{SynthStrip} brain mask~\cite{synthstrip}, as \texttt{FAE}, \texttt{DISYRE} and \texttt{dmprior} are reported. Specificity is measured on healthy subjects (\cref{sec:specificity}).

\noindent \textbf{Downstream evaluation}. Super-resolution is scored on ATLAS, acquired-isotropic BraTS \Tone{} and \Ttwo{} and WMH \FLAIR{}, degraded threefold (\cref{tab:sr}). Registration is scored on ATLAS, BraTS and WMH to \texttt{MNI152}~\cite{fonov2011} (\cref{tab:reg}; matched-displacement control in \appref{tab:reg_lesion}{sec:suppl_registration}).

\subsection{Baselines}

\noindent \textbf{Detection}. We compare to \texttt{FAE} and \texttt{DISYRE} published by the benchmark~\cite{uadbench} as contrast-specific models. The benchmark trained no \FLAIR{} model (``--'' in \cref{tab:main}). \texttt{UNA}~\cite{una} and \texttt{PEPSI}~\cite{pepsi} train on real pathological data and are not compared. No model in \cref{tab:main} sees real pathology in training. The control, the same UNet trained on healthy label maps only, with the tissue-class statistic as readout, is the ablation of the pathology itself (\cref{tab:main}). The data regime is ablated likewise: the same network trained on \texttt{DISYRE}'s pasted anomalies or on \texttt{FAE}'s healthy scans (\cref{sec:suppl_ablation}). Contrast-agnostic \texttt{dmprior}~\cite{dgp} runs with its multicontrast prior at default $\tau=0.05$, input minus refinement as the deviation map, with joint super-resolution and refinement on anisotropic scans.

\noindent \textbf{Super-resolution}. \texttt{dmprior}'s super-resolution task with a contrast-specific prior~\cite{dgp} serves as a reference trained on healthy data only.

\noindent \textbf{Registration}. \texttt{SynthMorph} denotes the published pretrained model~\cite{synthmorph,synthmorph_joint}, 
registered from the same affine. The same network trained under its protocol on our pool without pathology is the ablation of \cref{sec:suppl_downstream}. 

\subsection{Training}
\label{sec:training}

\noindent \textbf{Detection.} Trained from random weights for 500k steps with Adam at learning rate $10^{-4}$, two seeds; the last 100k steps decay the learning rate by cosine to $0.1\times$ and keep an exponential moving average of the weights (decay 0.9999), from which the reported model is read. A step holds the control rendering of one draw and, on half of the steps, its pathological rendering in the same batch (\cref{sec:loss}); the pathology-free control omits the second rendering and is otherwise identical. \textbf{Super-resolution.} The pathology-trained backbone at 300k steps, fine-tuned for 200k steps on healthy renderings with the calibrated residual head of \cref{sec:tasks} on the tricubic interpolation of the degraded draw; its control is the same backbone trained without pathology, under the identical fine-tune. \textbf{Registration.} The same backbone fine-tuned for 100k steps with the velocity head of \cref{sec:tasks}, the moving image pathological on half of the steps; its control is the same backbone trained without pathology under the identical fine-tune without pathology randomization. Every reported model is one set of weights: the EMA weights at 500k for detection and the final checkpoint of each fine-tune. Hyperparameters and ablations are in \cref{sec:suppl_training,sec:suppl_ablation}.

\section{Results}
\label{sec:results}

\begin{figure*}[t]
  \centering
  \IfFileExists{figs/sr_combined.png}{\includegraphics[width=\linewidth,height=0.42\textheight,keepaspectratio]{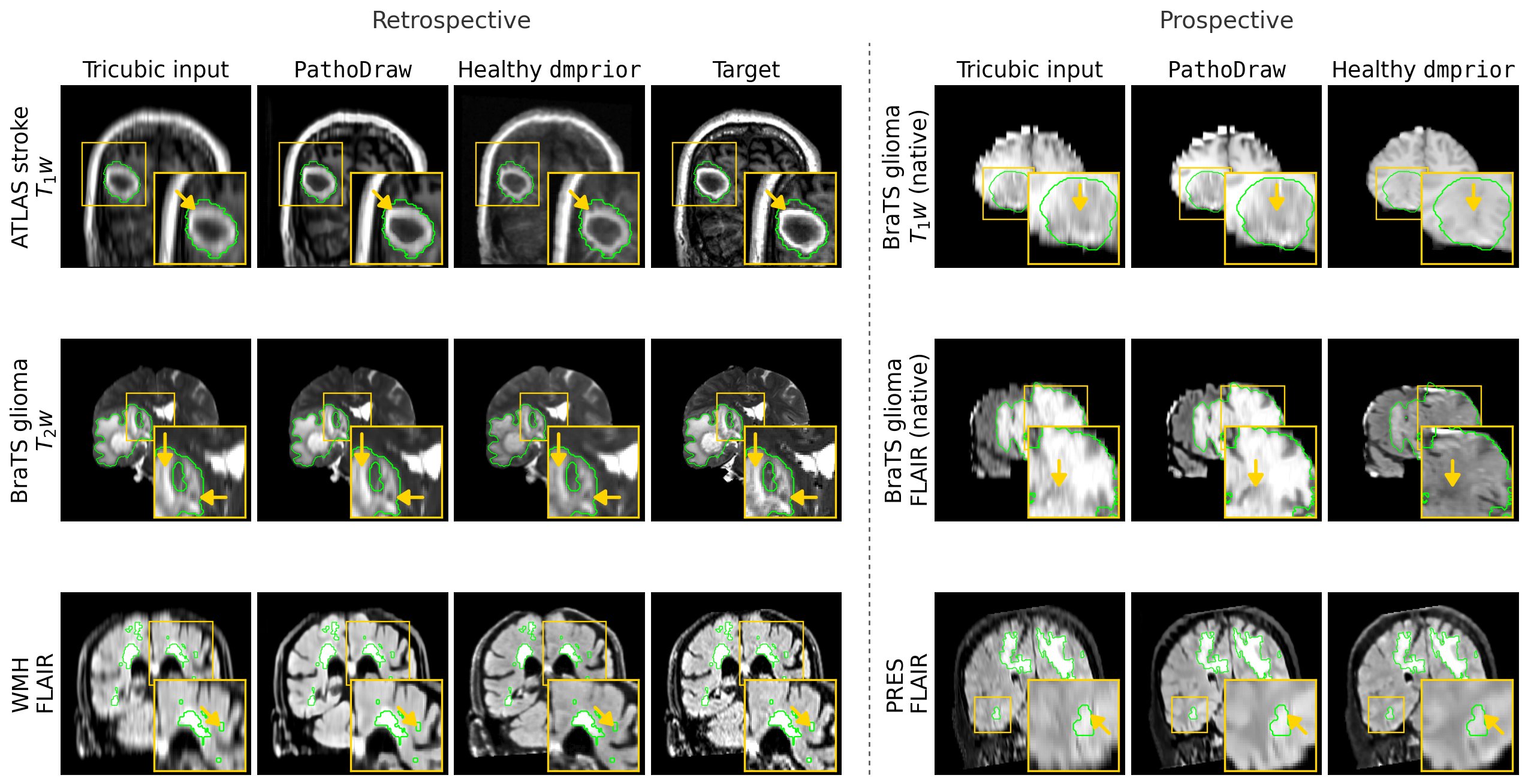}}{\fbox{combined super-resolution figure pending}}
  \caption{\textbf{Super-resolution}. Coronal slices with lesion outlines in green. Each row shows the interpolated input, \texttt{PathoDraw} (detector-modulated, as in \cref{tab:sr}), the healthy diffusion prior \texttt{dmprior}~\cite{dgp} at its contrast-specific super-resolution task, and a callout (yellow) on the lesion; the retrospective half (left of the dashed line, degraded slices, the clinical regime of the native thick-slice acquisitions on the right) adds the acquisition as target. Arrows point inside the lesion at the detail under comparison, where the healthy prior erodes or obscures pathology. The white matter hyperintensity row is degraded from its own $3\,$mm acquisition, which is its target.}
  \label{fig:sr}
\end{figure*}
\begin{figure*}[t]
  \centering
  \includegraphics[width=\linewidth,height=0.65\textheight,keepaspectratio]{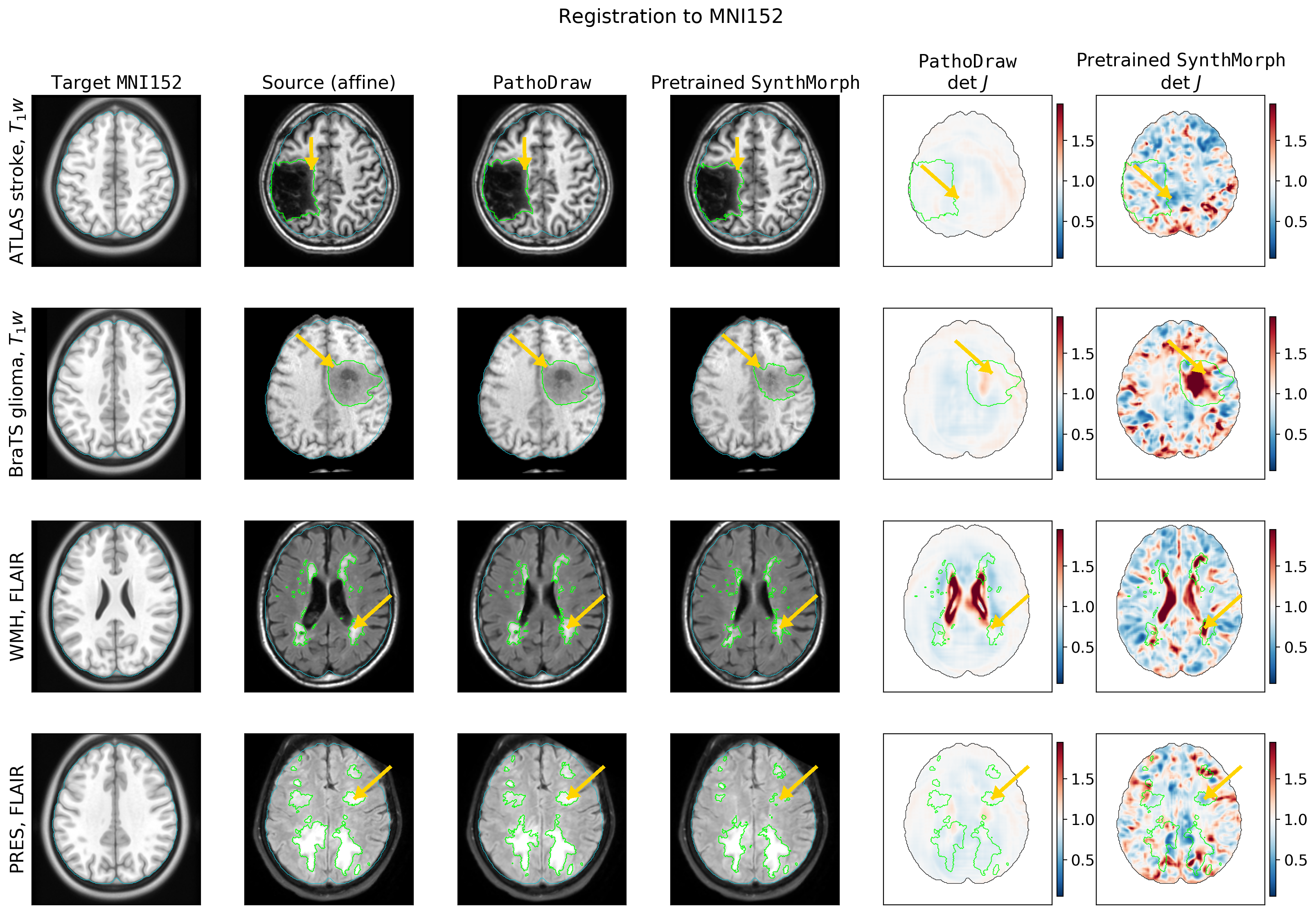}
  \caption{\textbf{Registration to \texttt{MNI152}}. Four cases at axial lesion slice with the template, source, warps and Jacobian determinants of our field and \texttt{SynthMorph} at default regularization (\cref{tab:reg}). Template brain mask in cyan, yellow arrows show lesion distortion in \texttt{SynthMorph}.}
  \label{fig:reg_mni}
\end{figure*}
\subsection{Zero-shot detection benchmark}

Our model exceeds or matches benchmarked approaches on four datasets (\cref{tab:main}, \cref{fig:teaser}). $\lceil$Dice$\rceil$ is the benchmark's oracle quantity, the same for every column; operating points are set on healthy scans in \cref{sec:specificity}.
Lesion size and contrast bound absolute Dice with median volumes of $5$\,mL on ATLAS, $15$\,mL on WMH and $89$\,mL on BraTS. WMH masks are drawn on \FLAIR{}; on \Tone{} those lesions are difficult to visualize. On ADNI3, a neurodegenerative cohort, the anatomy head agrees with the control's for cognitively normal and impaired subjects alike (\cref{sec:suppl_adni}).

\begin{table}[t]
  \caption{Zero-shot detection on the UAD benchmark~\cite{uadbench}, with 
   $\lceil$Dice$\rceil$~
  (\cref{sec:experiments}), %
  ``--'' when not evaluated. 
  \texttt{PathoDraw}, the control (trained without pathology randomization, \cref{sec:suppl_ablation}) and \texttt{dmprior} on the benchmark (%
  \cref{sec:suppl_benchmark}), PRES on every case; paired $p$-values in \cref{tab:pvalues}. Bold: the row maximum and any value inside \texttt{PathoDraw}'s 95\,\% bootstrap interval of it (\cref{sec:suppl_pooled}).}
  \label{tab:main}
  \centering\scriptsize\setlength{\tabcolsep}{1pt}
  \begin{tabular*}{\linewidth}{@{\extracolsep{\fill}}lccccc@{}}
    \toprule
    Dataset & \texttt{PathoDraw} & \begin{tabular}[c]{@{}c@{}}Control\\(no pathology)\end{tabular} & \begin{tabular}[c]{@{}c@{}}\texttt{dmprior}\\(no pathology)\end{tabular} & \texttt{FAE} & \texttt{DISYRE} \\
    \midrule
    \begin{tabular}[c]{@{}l@{}}Single model
    \end{tabular} & \checkmark & \checkmark & \checkmark & $\times$ & $\times$ \\
    ATLAS-O \Tone{} & \textbf{0.369} & 0.184 & 0.107 & 0.167 & \textbf{0.386} \\
    BraTS \Tone{}   & \textbf{0.537} & 0.228 & 0.129 & 0.437 & 0.274 \\
    BraTS \FLAIR{} & \textbf{0.622} & 0.494 & 0.214 & -- & -- \\
    BraTS \Ttwo{}   & \textbf{0.657} & 0.462 & 0.185 & 0.467 & \textbf{0.646} \\
    WMH \FLAIR{}   & \textbf{0.259} & \textbf{0.255} & 0.039 & -- & -- \\
    WMH \Tone{}     & \textbf{0.229} & 0.072 & 0.021 & 0.032 & 0.130 \\
    PRES \FLAIR{}  & \textbf{0.225} & 0.205 & 0.123 & -- & -- \\
    \bottomrule
  \end{tabular*}
\end{table}

\subsection{False-positive analysis with healthy subjects}
\begin{table}[t]
  \centering\scriptsize\setlength{\tabcolsep}{2.5pt}
  \caption{\textbf{Super-resolution}. Mean PSNR margin (dB) of detector-modulated \texttt{PathoDraw} over the tricubic input, the control trained without pathology randomization, and \texttt{dmprior}; mean $\pm$ standard deviation over seeds; modulation on degraded inputs (BraTS: \FLAIR{}); \texttt{dmprior} on its completed cases (170 / 550 / 358 / 55); bold: significant gain (\cref{tab:pvalues}).}
  \label{tab:sr}
  \begin{tabular}{@{}l l c c c@{}}
    \toprule
    & & \begin{tabular}[c]{@{}c@{}}Tricubic\\interpolation\end{tabular} & \begin{tabular}[c]{@{}c@{}}Control\\(no pathology)\end{tabular} & \begin{tabular}[c]{@{}c@{}}\texttt{dmprior}\\(no pathology)\end{tabular} \\
    \midrule
    ATLAS-O \Tone{} (210) & Brain  & $\mathbf{+0.96{\pm}0.03}$ & $\mathbf{+0.32{\pm}0.18}$ & $\mathbf{+4.74{\pm}0.03}$ \\
                          & Lesion & $\mathbf{+0.25{\pm}0.03}$ & $\mathbf{+0.60{\pm}0.28}$ & $\mathbf{+5.76{\pm}0.05}$ \\
    BraTS \Tone{} (690)   & Brain  & $\mathbf{+0.79{\pm}0.17}$ & $\mathbf{+0.47{\pm}0.47}$ & $\mathbf{+5.20{\pm}0.16}$ \\
                          & Lesion & $\mathbf{+0.24{\pm}0.03}$ & $\mathbf{+0.74{\pm}0.49}$ & $\mathbf{+7.47{\pm}0.03}$ \\
    BraTS \Ttwo{} (358)   & Brain  & $\mathbf{+1.09{\pm}0.08}$ & $-0.18{\pm}0.10$ & $\mathbf{+3.52{\pm}0.08}$ \\
                          & Lesion & $\mathbf{+0.44{\pm}0.07}$ & $+0.10{\pm}0.22$ & $\mathbf{+7.55{\pm}0.07}$ \\
    WMH \FLAIR{} (60)     & Brain  & $\mathbf{+0.68{\pm}0.02}$ & $-0.12{\pm}0.06$ & $\mathbf{+2.03{\pm}0.02}$ \\
                          & Lesion & $\mathbf{+1.48{\pm}0.22}$ & $-0.02{\pm}0.41$ & $\mathbf{+8.55{\pm}0.22}$ \\
    \bottomrule
  \end{tabular}
\end{table}

\label{sec:specificity}

\noindent 
We measure the deviation map on healthy scans, with thresholds set on HCP-Aging~\cite{hcpaging}~\cite{uadbench}. \texttt{PathoDraw} is comparable to the published models at matched specificity (\appref{tab:matched_fpr}{sec:suppl_fpr}). We analyze false positives in pathological datasets in \cref{sec:suppl_failure}.

\subsection{Task generalization}

\label{sec:results-downstream}

\noindent \textbf{Super-resolution}. Each acquisition is degraded along its slice axis, super-resolved by \texttt{PathoDraw} and \texttt{dmprior} (\cref{fig:sr}) and scored with peak signal-to-noise ratio (PSNR). The residual is scaled by $1-w$, the deviation complement on the degraded input (\cref{sec:suppl_sr}). \texttt{PathoDraw} improves tricubic input over the brain and lesion. Over the control trained without pathology randomization it is ahead on \Tone{} and level in the lesion on \Ttwo{} and \FLAIR{} (\cref{tab:sr}). Pathology does not conform to the piecewise-constant label model the renderer assumes, so the correction is scaled according to the deviation map, which bounds the margins. Despite learning exclusively from synthetic tissue rendered from piecewise-constant labels, the network still learns a mapping from interpolated low-resolution to high-resolution images. %

\noindent \textbf{Registration.} Against pretrained \texttt{SynthMorph} at its default regularization our field preserves the lesion, measured as its overlap with its affine source (\cref{tab:reg}; \appref{tab:reg_pretrained}{sec:suppl_pretrained}). The control (no pathology) preserves less of the lesion on every dataset. As smaller fields favor lesion Dice, \cref{sec:suppl_registration} repeats the comparison at matched mean displacement. When comparable, \texttt{PathoDraw} offers a better-conditioned Jacobian. \Cref{fig:reg_mni} shows lesion preservation where \texttt{SynthMorph} distorts it.

\begin{table}[t]
  \centering\footnotesize\setlength{\tabcolsep}{1.5pt}
  \caption{\textbf{Lesion Dice through the warp}, mean $\pm$ standard deviation over two seeds.}
  \label{tab:reg}
  \begin{tabular}{@{}l ccc@{}}
    \toprule
    & ATLAS-O & BraTS & WMH \\
    \midrule
    \texttt{PathoDraw} & $\mathbf{0.757{\pm}0.019}$ & $\mathbf{0.925{\pm}0.011}$ & $\mathbf{0.564{\pm}0.037}$ \\
    Control (no pathology) & $0.710{\pm}0.025$ & $0.911{\pm}0.013$ & $0.502{\pm}0.021$ \\
    \texttt{SynthMorph} & $0.644$ & $0.782$ & $0.384$ \\
    \bottomrule
  \end{tabular}
\end{table}

\section{Discussion}
\label{sec:discussion}

\noindent Our generative model of pathology randomization over healthy label maps
suffices for a single contrast-agnostic network robust to a variety of diseases zero-shot. Despite training purely on synthetic data, real lesions lie within the generator's shape manifold at high recall $0.86$--$0.93$ %
(\cref{sec:coverage}). The modes of pathology used in randomization may be broadened to improve the estimated manifold.

\smallskip\noindent\textbf{Limitations.}\label{sec:limitations}
Synthetic pathology is randomized over shape, placement and intensity rather than disease, so detectable conditions must lie on this manifold. 

\smallskip\noindent\textbf{Conclusion.}\label{sec:conclusion} We demonstrate our model trained with pathology randomization can detect stroke, glioma and white matter hyperintensities zero-shot, on public data and archived clinical scans. Our backbone extends to super-resolution and registration, and to any task whose target the forward model supplies. We release \texttt{PathoDraw} to expand the utility of uncurated clinical data and accelerate disease decoding in neuroimaging.

\clearpage

{
    \small
    \bibliographystyle{ieeenat_fullname}
    \bibliography{main}
}

\clearpage
\setcounter{table}{0}\renewcommand{\thetable}{S\arabic{table}}\renewcommand{\theHtable}{S\arabic{table}}
\setcounter{figure}{0}\renewcommand{\thefigure}{S\arabic{figure}}\renewcommand{\theHfigure}{S\arabic{figure}}
\setcounter{equation}{0}\renewcommand{\theequation}{S\arabic{equation}}\renewcommand{\theHequation}{S\arabic{equation}}
\maketitlesupplementary
\renewcommand{\topfraction}{0.95}\renewcommand{\bottomfraction}{0.95}\renewcommand{\textfraction}{0.05}\renewcommand{\floatpagefraction}{0.85}
\renewcommand{\dbltopfraction}{0.95}\renewcommand{\dblfloatpagefraction}{0.85}\setcounter{topnumber}{4}\setcounter{bottomnumber}{3}\setcounter{totalnumber}{8}\setcounter{dbltopnumber}{4}
\setcounter{section}{0}
\renewcommand{\thesection}{\Alph{section}}
\renewcommand{\theHsection}{appendix.\Alph{section}}\renewcommand{\theHsubsection}{\theHsection.\arabic{subsection}}\renewcommand{\theHsubsubsection}{\theHsubsection.\arabic{subsubsection}}

\section{Generator}
\label[appendix]{sec:suppl_generator}

\Cref{tab:generator} contains every range of the generator parameters and their sources, and the mode $\to$ label mapping. Below we describe the Rose criterion sampler $k\sim\mathcal{U}[1,2.5]$ 
and mass effects.

\begin{table*}[p]   %
  \caption{Generator summary: the conditional pathology distribution $p(\mathcal{P}\mid A)$ and the acquisition $p(\theta)$ of the reported model (\cref{sec:method}). Every entry is a fixed range or probability; no value is set on a real image. Modes are drawn as Bernoulli trials: one primary mode, then each other mode adds a second pathology independently with probability 0.15. Label ids follow the 0--56 scheme of the training label maps; $\sigma(\cdot)$ is the logistic function. Marks on the values: $^{\ast}$ design choice; $^{\dagger}$ literature value; $^{\ddagger}$ \texttt{SynthSeg} default.}
  \label{tab:generator}
  \centering\footnotesize
  \setlength{\tabcolsep}{5pt}\renewcommand{\arraystretch}{1.15}
  \begin{tabular}{@{}lp{0.60\textwidth}p{0.17\textwidth}@{}}
    \toprule
    Component & Setting & Provenance \\
    \midrule
    Anatomy $A\sim p_{\mathrm{data}}$ & 1{,}033 healthy label maps (HCP 821, Chinese HCP 212)$^{\dagger}$, 56 classes$^{\ast}$; $160^3$ crop at a random center$^{\ddagger}$; rotation $\mathcal{U}[-15,15]^\circ$, shear $\mathcal{U}[-0.2,0.2]$, scale $1+\mathcal{U}[-0.2,0.2]$ per axis, nonlinear field with standard deviation $\mathcal{U}[0,4]$ voxels on a 3--6\,\% grid$^{\ddagger}$; left/right flip with label swap, $p=0.5^{\ddagger}$; extracerebral block removed, $p=0.5^{\ast}$ & \texttt{SynthSeg} \cite{synthseg}; HCP \cite{hcp}, Chinese HCP \cite{chinesehcp} \\
    Modes of $\mathcal{P}$ & primary mode: local 0.23, vascular-aligned 0.385, tract-aligned 0.385$^{\ast}$ (diffuse and atrophy modes off); each other mode adds a second pathology with $p=0.15^{\ast}$ (28\,\% of draws carry $\ge2$ modes; co-occurrence$^{\dagger}$) & \cite{vermeer2002,deleeuw2001} \\
    Mode $\to$ labels & local: any brain label present (uniform over labels)$^{\dagger}$, ROI bias $p=0.8^{\ast}$, confined to that label $p=0.5^{\ast}$, 1--3 foci with weights 0.85 / 0.12 / 0.03$^{\dagger}$; vascular-aligned: perforator branch (thalamus 7/24, caudate 8/25, putamen 9/26, pallidum 10/27) or cortical branch (cerebral white matter 1/18, cortex 2/19)$^{\dagger}$, $p=0.5^{\ast}$ each; tract-aligned: one of lateral and inferior-horn (3, 4 / 20, 21) or third and fourth (11, 12) ventricles ($p=0.75^{\ast}$) or all of them, with cerebral white matter (1/18)$^{\dagger}$ & \cite{bamford1991,djulejic2016,fazekas1987,larjavaara2007}; foci \cite{patil2012} \\
    Geometry & excursion sets of a Gaussian random field with a spectral-mixture power spectrum$^{\dagger}$: 1--9 zero-centered Gaussian components, bandwidths $\mathcal{U}[2/160, f_{\max}/2]$ cycles/voxel, $f_{\max}=0.15+0.2\,\sigma(\mathcal{N}(0,1))$ (local) or $0.1+0.2\,\sigma(\mathcal{N}(0,1))$ (vascular, tract)$^{\ast}$; anisotropic warp, elongation $1+1.5\,\sigma(\mathcal{N}(0,1))$ per axis, rotation $\le40^{\circ\ast}$; gated by a focal prior (local: Gaussian envelope, $\sigma=0.6\times\mathcal{U}[0.10,0.20]\times160$ voxels; vascular: blob for the perforator branch, wedge or column for the cortical branch; tract: periventricular white matter band, tangent slab or Gaussian envelope)$^{\ast}$; the field's top $0.2+0.7\,\sigma(\mathcal{N}(0,1))$ (local, vascular) or $0.15+0.7\,\sigma(\mathcal{N}(0,1))$ (tract) of the prior becomes lesion$^{\ast}$ (thresholded noise$^{\dagger}$); core/shell ring always for local and tract, never for vascular$^{\ast}$; three GMM intensity zones per lesion, cut by a second field with components centered at $\mathcal{U}[1/160, f_{\max}]$ and scale $\max(6, 0.3\,r)$ voxels ($r$ = lesion half-width), one zone when $0.3\,r<6^{\ast}$ & \cite{adler2007rfg,wilson2013sm}; thresholded noise \cite{draem,una} \\
    Mass effect & smoothed, normalized gradient of the lesion potential (blur $\sigma=\mathcal{U}[0.10,0.11]\times160$ voxels, displacement smoothed with $\sigma=\mathcal{U}[0.005,0.02]\times160$)$^{\ast}$; per mode (tumor and infarct mass effect$^{\dagger}$): local $p=0.8$, magnitude $2+6\,\sigma(\mathcal{N}(0,1))$, random sign; vascular $p=0.4$, $3\,(2\sigma(\mathcal{N}(0,1))-1)$; tract $p=0.2$, $1+3\,\sigma(\mathcal{N}(0,1))$, random sign$^{\ast}$; total clamped at 26.25 (fold-free)$^{\ast}$ & tumor \cite{prastawa2009,clatz2005}; infarct \cite{battey2014,ritzl2004} \\
    Rose criterion & criterion$^{\dagger}$ with $k\sim\mathcal{U}[1.0,2.5]^{\ast}$; $\sigma_{\mathrm{surround}}$ = standard deviation of the control rendering over a 3-voxel-wide band starting 3 voxels outside the lesion (\cref{sec:intensity}; the estimated noise standard deviation when the band has $<27$ voxels)$^{\ast}$; a lesion whose mean rendered change is below $k\sigma_{\mathrm{surround}}$ is rebuilt to it (floor), one above $2k\sigma_{\mathrm{surround}}$ is scaled down to it (ceiling), so every drawn lesion satisfies the criterion against its surround; the geometric lesion label is kept in full$^{\ast}$ & \cite{rose1948} \\
    Sign & lesions rebuilt at the floor: $P(\text{hyperintense to the surround})=0.15$ where white matter renders brighter than cortex, 0.85 otherwise$^{\dagger}$ (flipped when only the other side has $k\sigma_{\mathrm{surround}}$ of headroom)$^{\ast}$; lesions above the floor as drawn: multi-zone lesions: one sign per lesion, $p=0.5^{\ast}$, every zone mean re-anchored to that side of the host class mean with its drawn magnitude, zone standard deviation $\le2\times$ host standard deviation$^{\ast}$ & \cite{bottomley1987,ginat2012} \\
    Composition & with $p=0.5$ / 0.5 / 0.3 (vascular / local / tract)$^{\ast}$ each zone renders at $\mu=(1-f)\mu_{\mathrm{host}}+f\mu_{\mathrm{mod}}$ with the host standard deviation (host = the label the lesion overwrites, modulating class = CSF or GM)$^{\dagger}$: vascular core $\to$ CSF $f\in[0.7,0.95]$, shell $\to$ GM $[0.3,1.0]$; local core $\to$ CSF $[0.5,0.95]$, shell $\to$ GM $[0.1,0.5]$; tract $\to$ GM $[0.3,1.0]$ (classes$^{\dagger}$, ranges$^{\ast}$); core = erosion to 15--50\,\%$^{\ast}$; applied only if $f\,|\mu_{\mathrm{host}}-\mu_{\mathrm{mod}}|\ge\sigma_{\mathrm{host}}$, otherwise the Rose-gated zone above$^{\ast}$; exempt from floor and ceiling$^{\ast}$ & \cite{strive,allen2012,upadhyay2011} \\
    Rendering $\theta$ & class means $\mathcal{U}[25,225]$, standard deviation $\mathcal{U}[5,25]$, background mean zeroed with $p=0.5^{\ddagger}$; dark CSF on half the draws$^{\ast}$, $\mu_{\mathrm{CSF}}=\mu_{\mathrm{WM}}\,\mathcal{U}[0.10,0.40]^{\dagger}$; partial-volume ramps$^{\dagger}$; synthesis in white matter units$^{\ast}$; gamma $\exp(0.1\,\mathcal{N}(0,1))^{\ast}$; multiplicative bias field, log-standard deviation $\mathcal{U}[0.1,0.6]$ on a $\mathcal{U}[0.02,0.04]\times160$ grid$^{\ddagger}$; resolution isotropic $\mathcal{U}[1,4]$\,mm or one axis $\mathcal{U}[1,8]$\,mm (equiprobable), slice thickness = resolution, blur standard deviation $(0.85+0.3\,\mathcal{U}[0,1])\ln5/\pi\times$ thickness$^{\ddagger}$; acquisition noise standard deviation $\mathcal{U}[0.05,0.15]\,\mu_{\mathrm{WM}}{}^{\ast}$; per-image max normalization$^{\ddagger}$; each draw trains three steps$^{\dagger}$: as rendered, then twice re-perturbed with extra noise standard deviation $\mathcal{U}[0.01,0.05]$ and gamma $\mathcal{U}[0.85,1.15]^{\ast}$ & \texttt{SynthSeg} \cite{synthseg}; partial volume \cite{vanleemput2003}; slice profile \cite{billot2020pv}; bias \cite{sled1998}; noise \cite{gudbjartsson1995}; dark CSF \cite{hajnal1992,rooney2007,wansapura1999,mugler1990}; three steps per draw \cite{choi2019echoing} \\
    Shared per step & subject, crop, warps, flip, class table, synthesis noise, gamma, bias, resolution, noise: identical for the control and the pathological rendering; only the label map (lesion, mass effect) and the visibility correction differ (common random numbers$^{\dagger}$) & \cite{glasserman1992crn} \\
    Target $y$ & $\tanh(\Delta I/s)$, $s=\wp_{99}-\wp_{1}$ of the pathological rendering$^{\dagger}$, on the lesion support with sign-matched floor $C=0.18^{\ast}$, zero elsewhere and zero on the control rendering$^{\ast}$ & \cref{sec:method}; intensity frame \cite{synthseg,synthsr,synthmorph} \\
    \bottomrule
  \end{tabular}
\end{table*}

\subsection{Derivation of the joint distribution}
\label[appendix]{sec:suppl_joint}
The factorization of \cref{eq:gen} follows from the chain rule and two independence assumptions that the generator satisfies by construction. By the definition of conditional probability,
  \begin{equation}
    p(x\mid A,\mathcal{P},\theta)=\frac{p(x,A,\mathcal{P},\theta)}{p(A,\mathcal{P},\theta)},
    \label{eq:suppl_chain}
  \end{equation}
  \begin{equation}
    p(\mathcal{P}\mid A,\theta)=\frac{p(A,\mathcal{P},\theta)}{p(A,\theta)},
    \label{eq:suppl_chain2}
  \end{equation}
and $p(A,\theta)=p(\theta)\,p(A\mid\theta)$. The acquisition is drawn without reference to the anatomy, $A\perp\theta$, so $p(A,\theta)=p(A)\,p(\theta)$ and \cref{eq:suppl_chain2} becomes
  \begin{equation}
    p(\mathcal{P}\mid A,\theta)=\frac{p(A,\mathcal{P},\theta)}{p(A)\,p(\theta)}.
    \label{eq:suppl_chain3}
  \end{equation}
Substituting \cref{eq:suppl_chain3} into \cref{eq:suppl_chain},
  \begin{equation}
    p(x\mid A,\mathcal{P},\theta)=\frac{p(x,A,\mathcal{P},\theta)}{p(\mathcal{P}\mid A,\theta)\,p(A)\,p(\theta)}.
  \end{equation}
The pathology is drawn from the anatomy alone, before any intensity exists (\cref{sec:prior-labels}), so $\mathcal{P}\perp\theta\mid A$ and $p(\mathcal{P}\mid A,\theta)=p(\mathcal{P}\mid A)$, which gives
  \begin{equation}
    p(x,A,\mathcal{P},\theta)=p(x\mid A,\mathcal{P},\theta)\,p(\mathcal{P}\mid A)\,p(A)\,p(\theta),
  \end{equation}
\cref{eq:gen}. Every lesion quantity that does depend on $\theta$, the composed class means and the Rose-criterion rebuild of \cref{sec:intensity}, is part of the rendering and therefore of $p(x\mid A,\mathcal{P},\theta)$, not of $p(\mathcal{P}\mid A)$.

\subsection{Pathology randomization}
\label[appendix]{sec:suppl_prior}

\begin{figure*}[t]
  \centering
  \IfFileExists{figs/modes.png}{\includegraphics[width=\linewidth,height=0.60\textheight,keepaspectratio]{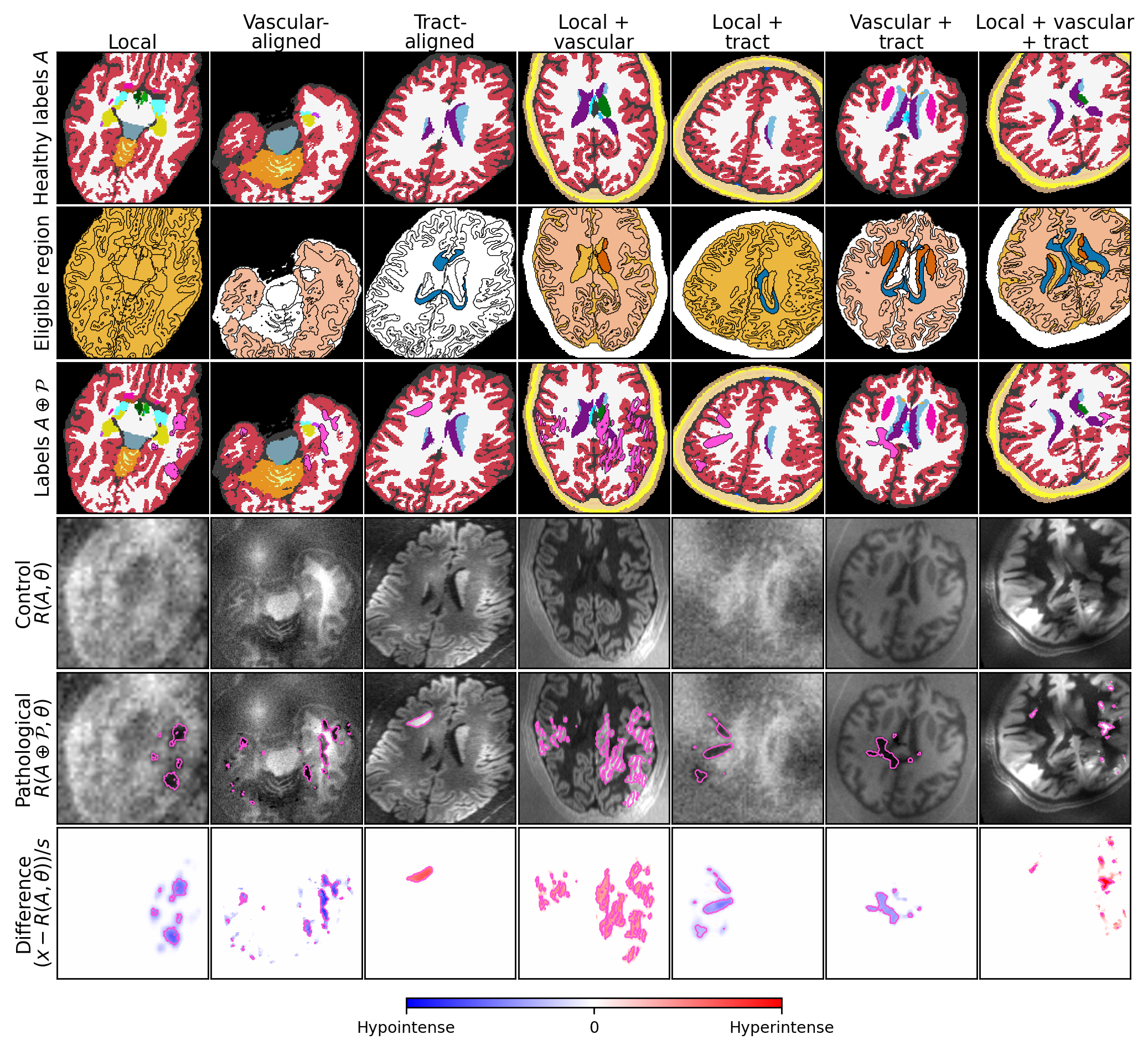}}{\fbox{modes figure pending}}
  \caption{Pathology modes of $p(\mathcal{P}\mid A)$: one generator draw per column (local, vascular-aligned, tract-aligned, and their combinations) and one stage per row (healthy labels $A$, the mode's eligible region, the injected draw in pink, the control and pathological renderings under one acquisition draw, and their difference).}
  \label{fig:modes}
\end{figure*}

\noindent \textbf{Modes}. Drawn modes consist of tract-aligned (periventricular: seeded from the ventricles into the white matter), vascular-aligned (deep gray nuclei, or cerebral white matter and cortex), and local (any brain label) \cite{strive}. Within these modes, pathology geometry is conditioned on the anatomy through each mode's eligible region and envelope (\cref{eq:support}, mode $\to$ labels row of \cref{tab:generator}), shown per mode in \cref{fig:modes}. Mass effects are computed by a smoothed gradient of the lesion potential. Each draw is a chain of independent trials, beginning with a primary mode $m\sim\operatorname{Cat}(0.23,\,0.385,\,0.385)$ over (local, vascular-aligned, tract-aligned). Every other mode is added by an independent $\operatorname{Bern}(0.15)$ draw. There may be one to three foci in the local mode, %
a perforator or cortical branch ($p=0.5$ each) for the vascular-aligned mode and a periventricular band seeded from one ventricle group for the tract-aligned mode, the lesion inside any envelope being the excursion set of \cref{eq:support}, which can fall into several components in every mode, so no mode is limited to one focus. Each mode adds mass effect with probability $0.8$ / $0.4$ / $0.2$ per mode (local / vascular-aligned / tract-aligned). The composition rendering below is applied with probability $0.5$ / $0.5$ / $0.3$ per mode. \Cref{tab:generator} lists every probability and range.

\noindent \textbf{Intensity}. Every label renders as a Gaussian $\mathcal{N}(\mu_l,\sigma_l^{2})$ (\cref{eq:render}), including lesion labels. With its mode's composition probability ($0.5$ / $0.5$ / $0.3$ above) a lesion is rendered by composition rather than with a label mean of its own. Otherwise, a lesion renders as an ordinary label under the Rose criterion below. A zone of the composed lesion, its core or its shell, retains the host's $\sigma$ and takes the mean $\mu=(1-f)\mu_{\mathrm{host}}+f\mu_{\mathrm{mod}}$, where the host is the healthy label the zone overwrites and the modulating class a second healthy label whose mean modulates the zone's. 
The acquisition draw $\theta$ already holds one mean $\mu_l$ per label (\cref{eq:render}), and the zone reads the host's and the modulating label's from it. The modulating label is fixed by the zone, CSF for a core (cavitation, encephalomalacia, necrosis) \cite{strive,allen2012,upadhyay2011} and gray matter for a shell or a tract-aligned lesion (gliosis, WMH) \cite{strive}. Only the fraction $f$ is drawn, per zone, from the ranges in \cref{tab:generator}. A zone whose composition is invisible against its host ($f\,|\mu_{\mathrm{host}}-\mu_{\mathrm{mod}}|<\sigma_{\mathrm{host}}$) falls back to the ordinary path. As both means belong to the same $\theta$, the zone is darker or brighter than its host exactly as CSF or gray matter is in that draw, so the lesion polarity follows the drawn contrast, and the composition is contrast-agnostic by construction.

\noindent \textbf{Rose criterion} \cite{rose1948}\textbf{.} Every drawn lesion must be visually perceptible with respect to its surrounding tissue as defined by contrast-to-noise ratio $\mathrm{CNR}=|\overline{\Delta I}_L|/\sigma_{\mathrm{surround}}\ge k$, where $\overline{\Delta I}_L$ is the lesion's mean rendered contrast and $\sigma_{\mathrm{surround}}$ the standard deviation of the control rendering in a 3-voxel band around it. To randomize contrast sufficiently the required CNR itself varies, $k\sim\mathcal{U}[1,2.5]$, and the renderer raises a lesion below $k$ to it and lowers one above $2k$ to that ceiling (\cref{tab:generator}). Composition lesions are exempt from this floor and instead gated at $k=1$ against their host class. \Cref{fig:modes} shows one draw per mode and combination and \cref{fig:flow} follows one draw through every stage of pathology randomization.

\section{Region-target floor}
\label[appendix]{sec:suppl_rose}
The Rose criterion admits a lesion whose mean contrast is at least $k\sigma_{\mathrm{surround}}$ with $k\ge1$ (\cref{sec:intensity}). The smallest deviation the renderer calls visible is therefore $\sigma_{\mathrm{surround}}$ ($k=1$), which in target units (\cref{sec:targets}) is $\tanh(\sigma_{\mathrm{surround}}/s)$. Its median over 200 pregenerated draws is 0.184 on one set and 0.182 on an independent second set, hence
\begin{equation}
  C=\operatorname*{med}_{\text{draws}}\,\tanh\!\big(\sigma_{\mathrm{surround}}/s\big)\approx0.18.
  \label{eq:floor}
\end{equation}
 A lesion voxel whose rendered contrast falls below this floor is still labeled lesion by the pathology model, so its target is raised to the region target $C$ with the voxel's own sign, or the lesion's mean sign where $\Delta I\approx0$.

\section{Ablation details}
\label[appendix]{sec:suppl_ablation}

\noindent \textbf{Development protocol.} Attention gates on the skip connections \cite{attnunet,olchanyi2026brainstem} and dark-CSF rendering \cite{hajnal1992,rooney2007,wansapura1999,mugler1990} were adopted using robustness to distribution shift
in synthetic data alone. Distribution shift was simulated on a validation set re-rendered with every appearance and nuisance range of the generator widened to twice the training ranges. The model with both components is above the control model in both tested seeds (shell AUC $+0.016$ $[+0.005, +0.026]$ and $+0.059$ $[+0.047, +0.072]$ over 566 cases with a visible lesion). Neither component alone separates. \Cref{tab:ablation} gives the per-dataset gains on the real data. 

\noindent \textbf{Data-regime ablation.} As the pathology randomization generator is our contribution, the data regime is ablated on the same architecture, loss, healthy pool and step budget, every model trained to 500k and read out with batch-normalization statistics of the input scan (\cref{tab:regime}). This ablation contains (a) synthetic renderings with pathology randomization, the reported model, (b) the \texttt{DISYRE} analog, real healthy scans with a foreign patch of another real scan pasted in as the anomaly \cite{uadbench}, read through the deviation head, (c) the \texttt{FAE} analog, real healthy scans with no anomaly, read through the tissue-class statistic, the readout of the control of \cref{tab:main}, and through its own head, which has no deviation to learn. The two real-image regimes keep each scan's acquired contrast and skip contrast randomization and dark-CSF rendering; only the acquisition stage after rendering still applies, so neither sees a \FLAIR{} scan, the pool's real contrasts being \Tone{} and \Ttwo{}. Pathology randomization leads on ATLAS and WMH \Tone{} against both real regimes and is within $0.02$ of the healthy-scan statistic on the two \FLAIR{} datasets, ahead on BraTS and behind on WMH. The pasted patches lead on BraTS \Tone{} and \Ttwo{}. We expect this to follow from tumors consisting of abnormal growth of the tissue around them. On \Tone{} and \Ttwo{} contrasts a tumor presents largely as near-normal tissue in the wrong place with mass effect, within the distribution of a pasted patch of real brain. Conversely, a stroke cavity, \FLAIR{} edema and white matter hyperintensities are intensity deviations outside the healthy tissue distribution. Removing pathology randomization from the synthetic regime gives the control model of \cref{tab:main} when read through the robust tissue-class statistic; read through its deviation head it is near chance, as regime (c) is, since there is no deviation to learn.

\begin{table*}[tbp]
  \centering\small\setlength{\tabcolsep}{5pt}
  \caption{\textbf{Data-regime ablation.} Ceiling Dice on every case, mean of two seeds (500k steps, EMA weights, 4\,mm median filter): our network and loss trained on (a) synthetic renderings with pathology randomization, the reported model, (b) the \texttt{DISYRE} analog, real healthy scans with pasted foreign patches, and (c) the \texttt{FAE} analog, real healthy scans alone, read through the tissue-class statistic or its own deviation head. Bold: column maximum.}
  \label{tab:regime}
  \begin{tabular}{@{}l cccccc@{}}
    \toprule
    Regime & ATLAS & BraTS \Tone{} & BraTS \FLAIR{} & BraTS \Ttwo{} & WMH \FLAIR{} & WMH \Tone{} \\
    \midrule
    (a) Synthetic, pathology randomization & \textbf{0.336} & 0.539 & \textbf{0.654} & 0.653 & 0.268 & \textbf{0.239} \\
    (b) Real scans, pasted patches & 0.261 & \textbf{0.585} & 0.547 & \textbf{0.676} & 0.115 & 0.075 \\
    (c) Real healthy scans, statistic readout & 0.165 & 0.246 & 0.631 & 0.544 & \textbf{0.288} & 0.118 \\
    (c) Real healthy scans, head readout & 0.032 & 0.128 & 0.120 & 0.120 & 0.038 & 0.036 \\
    \bottomrule
  \end{tabular}
\end{table*}

\begin{table*}[tbp]
  \caption{Ablation of dark CSF and attention on six datasets from a common backbone at 300k to 400k following distribution shift robustness in synthetic validation data.
  Reported as $\Delta\lceil$Dice$\rceil$ (median-filtered, the benchmark's convention) as mean $\pm$ standard deviation over the two seeds.}
  \label{tab:ablation}
  \centering\small\setlength{\tabcolsep}{4pt}   %
  \begin{tabular}{@{}lcccccc@{}}
    \toprule
    & ATLAS & BraTS \Tone{} & BraTS \FLAIR{} & BraTS \Ttwo{} & WMH \FLAIR{} & WMH \Tone{} \\
    \midrule
    + attention gates & $+0.018\pm0.019$ & $+0.012\pm0.002$ & $+0.071\pm0.032$ & $+0.019\pm0.016$ & $+0.056\pm0.064$ & $+0.054\pm0.037$ \\
    + dark-CSF rendering & $+0.029\pm0.020$ & $+0.087\pm0.024$ & $+0.095\pm0.047$ & $+0.062\pm0.054$ & $+0.127\pm0.050$ & $+0.005\pm0.036$ \\
    + both (reported) & $+0.039\pm0.026$ & $+0.029\pm0.009$ & $+0.147\pm0.059$ & $+0.063\pm0.036$ & $+0.067\pm0.046$ & $+0.040\pm0.049$ \\
    \bottomrule
  \end{tabular}
\end{table*}

\section{Robust tissue-class statistic}
\label[appendix]{sec:suppl_statistic}

\noindent \textbf{Robust tissue-class statistic.} The intensity $x_v$ of voxel $v$ is rescaled to $\tilde x_v=\mathrm{clip}\!\left((x_v-\mathrm{med}_{c(v)})/(1.4826\,\mathrm{MAD}_{c(v)}+\varepsilon),\,-10,\,10\right)$ according to robust statistics within its anatomy-head class $c(v)$. The scaling requires median $\mathrm{med}_{c(v)}$ and median absolute deviation (MAD) over the class's $c$ voxels $v$ in the acquisition. 
The robust tissue-class statistic requires no parameters and no gradients. For models trained without pathology randomization, the robust tissue-class statistic serves as a substitute deviation map.
The anatomy head never predicts a lesion class (its target is the healthy map under the lesion), so a lesion voxel is scored against its host tissue.

\section{Downstream heads}
\label[appendix]{sec:suppl_downstream}

\noindent \textbf{Controls.} The control of each task is the control model without pathology randomization, a detector trained for the same 300k steps with pathology randomization switched off, fine-tuned for the same steps on healthy renderings only. The anatomy pool, steps, seed and renderings are fixed so the only difference is lack of pathology randomization in backbone training data. For registration the reported arm itself adds the pathological rendering to the fine-tune on half of the steps (\cref{tab:reg_lesion}). We claim robustness under pathology, rather than accuracy over the pretrained \texttt{SynthMorph} model or \texttt{dmprior}.

\section{Loss}
\label[appendix]{sec:suppl_loss}

\noindent \textbf{Deviation loss.} Over the batch holding the control and the pathological image of one draw (\cref{sec:loss}),
  \begin{equation}
    \begin{aligned}
      \mathcal{L}_{\mathrm{dev}}&=\underbrace{\tfrac{1}{\Omega}\sum_{v}\omega_v\,|\hat r_v-y_v|}_{\mathcal{L}_{R(A\oplus\mathcal{P})}}+\underbrace{\tfrac{1}{|V|}\sum_{v}|\hat r^{0}_v|}_{\mathcal{L}_{R(A)}},\\
      \omega_v&=\begin{cases}|L^{c}|/|L| & v\in L,\\ 5 & v\in H,\\ 1 & \text{otherwise},\end{cases}
    \end{aligned}
    \label{eq:loss}
  \end{equation}
  with $\Omega=\sum_v\omega_v$. The first sum runs over the pathological rendering and the second over the control, where the deviation must vanish. The weights emphasize regions relevant in the detection task. The ratio $|L^{c}|/|L|$ balances the lesion against the far larger background, and $H$, the 5-voxel dilation of $L$ outside it, is the border band. The cross-entropy is likewise weighted $5$ inside the ROI (the healthy structures the lesion overlaps), the border-band and class weighting of~\cite{unet,dbce}, fixed a priori.

\section{Super-resolution}
\label[appendix]{sec:suppl_sr}
\noindent \textbf{Scoring} (\cref{tab:sr}). Each input is degraded along its slice axis: blurred with a Gaussian of standard deviation $\ln 5/\pi$ times the slice thickness, the slice-profile convention of the \texttt{Brain-ID}, \texttt{PEPSI} and \texttt{UNA} generator code~\cite{brainid,pepsi,una} (in training the same acquisition stage draws the degraded axis at random), sampled at three times the native slice spacing and restored to the native grid by tricubic interpolation, which is the network's input and the interpolation baseline. Input and target share one normalization constant and the input's $[\wp_1,\wp_{99}]$ window, the frame the network was trained in. The head predicts a residual and, next to it, a per-voxel amplitude in $(0,1)$ that scales it, trained jointly under the same loss; nothing but the input enters at inference. PSNR is read within the \texttt{SynthStrip}~\cite{synthstrip} brain mask and inside the dataset's lesion mask. The acquisition the output is scored against is never read, by the network or by the gate.

\subsection{Acquisitions}
\label[appendix]{sec:suppl_sr_acq}
\noindent \textbf{ATLAS-O.} The 210 \Tone{} scans of the benchmark's unseen-scanner subset, 1\,mm, degraded to 3\,mm; the lesion mask is the dataset's and the modulation map is the detector on the degraded \Tone{} itself.

\noindent \textbf{BraTS.} The template-registered 1\,mm volumes are resampled from acquisitions of unknown geometry, so automatic anisotropy detection on the acquired volumes (\cref{sec:suppl_aniso}; 690 of 1{,}249 near-isotropic) splits the super-resolution result in two: on the acquired-isotropic volumes the gain over the tricubic input is $+1.0$\,dB over the brain and $-0.1$ inside the tumor (unmodulated, one seed); on the 559 resampled volumes it is $-1.3$ and $-5.7$\,dB, where the tricubic input already scores $37.6$\,dB inside the tumor because the ``1\,mm'' target is itself an interpolation of a thick acquisition and every restored edge counts as error. The tables report the acquired-isotropic subset, 690 \Tone{}, 358 \Ttwo{} and 316 \FLAIR{} volumes by the same two-component split; the exclusion is by acquisition, stated once. For the BraTS \Tone{} and \Ttwo{} columns of \cref{tab:sr} the modulation map is the detector on the session's co-registered \FLAIR{}, degraded exactly like the input (3\,mm along the same axis), where the tumor is faint on \Tone{} (\cref{tab:main}); in \cref{tab:sr_contrast} it is the detector on each contrast's own degraded input. %

\begin{figure*}[t]
  \centering
  \includegraphics[width=\linewidth,height=0.32\textheight,keepaspectratio]{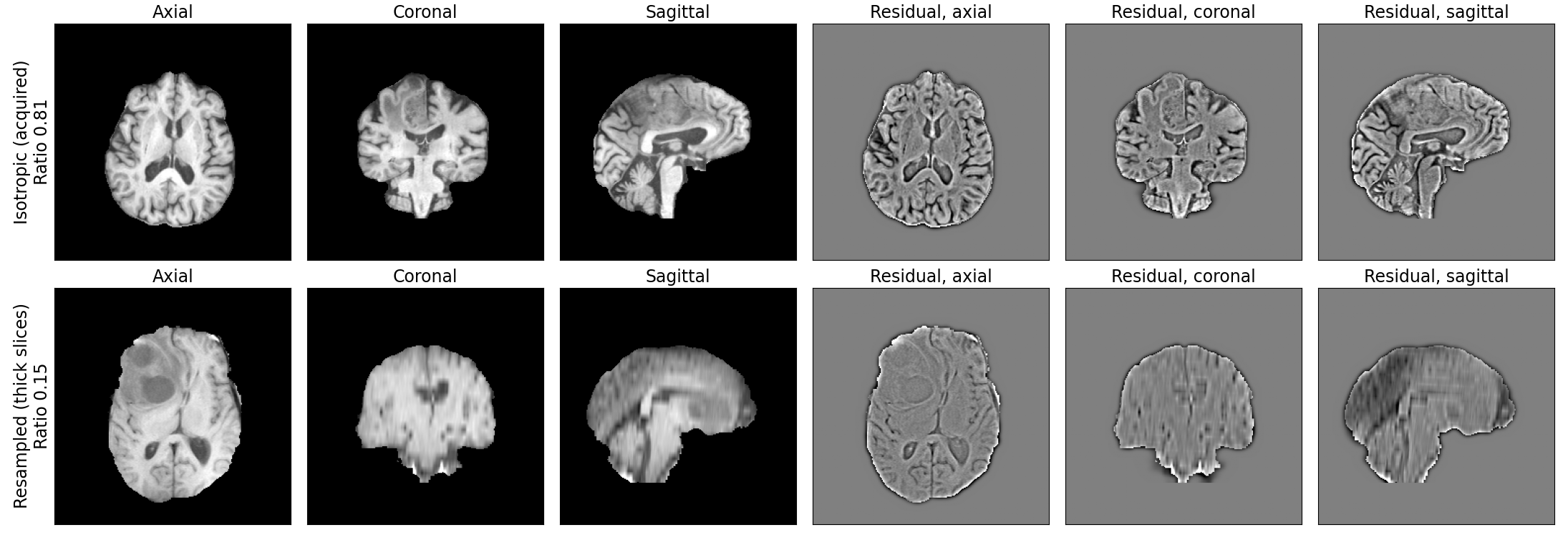}
  \caption{\textbf{Acquired-isotropic versus resampled BraTS volumes} (\cref{sec:suppl_aniso}). Left three columns: the \Tone{} volume in the three orthogonal planes; right three: its high-pass residual. Top: an acquired near-isotropic volume (ratio 0.81); bottom: a volume resampled from thick slices (0.15), smooth through-plane in the coronal and sagittal views.}
  \label{fig:brats_aniso}
\end{figure*}

\noindent \textbf{WMH.} Every site's \FLAIR{} is a 3\,mm axial acquisition, so it is degraded threefold in voxels (3 to 9\,mm) along its slice axis and scored against the acquisition on its own grid, never an interpolated target; the manual mask is drawn on that grid and the modulation map is the detector on the degraded \FLAIR{} itself. The 2D \Tone{} of the same subjects shares that grid. The 1\,mm 3D \Tone{} (59 subjects; one Utrecht subject at $0.89$\,mm in-plane excluded) is degraded 1 to 3\,mm with the \FLAIR{} mask rigidly resampled onto it (\texttt{mri\_coreg}, nearest).

\subsection{Automatic anisotropy detection}
\label[appendix]{sec:suppl_aniso}
Template registration to 1\,mm rotates the original slice direction off the grid axes and interpolates along all three, so axis-aligned tests such as per-axis gradient energy or spectra cannot detect a thick acquisition. The anisotropic point-spread function along the original slice direction, is unchanged however, which we measure rotation-invariantly. The high-pass residual $\tilde r$ of the volume (the image minus a Gaussian blur of 2 voxels) is taken inside the foreground eroded away from its edge, its gradient structure tensor $J=\sum_v \nabla \tilde r_v \nabla \tilde r_v^{\top}$ is accumulated over that region, and the statistic is the eigenvalue ratio $\lambda_{\min}/\lambda_{\max}$, which is 1 for an isotropic point-spread function and falls as the slices thicken. The residual is used rather than the image because anatomy has its own anisotropy, while noise does not, and interpolation correlates noise along the interpolated axis. The ratio is bimodal over the dataset (modes near 0.2 and 0.8, \cref{fig:brats_aniso}). A two-component Gaussian mixture fit to all 1{,}249 values, with no reference to labels or scores, assigns a volume to the near-isotropic component where its posterior exceeds one half: 690 \Tone{}, 358 \Ttwo{} and 316 \FLAIR{} volumes.

\subsection{Detector modulation}
\label[appendix]{sec:suppl_sr_gate}
\noindent \textbf{Detector modulation.} The residual is the high-frequency content the healthy prior restores. Inside pathology, the true content differs from a healthy prior's, and the residual is unreliable. The detector's deviation captures voxels where the healthy prior does not apply, and scales the residual toward the interpolant there. At inference the output is $\bar x+(1-w)\,f(\bar x)$ with $w=\min(1,|\delta|/\tau)$, $\delta\in[-1,1]$ the detector's signed deviation at the voxel and $\tau=0.4$, the value fixed a priori. The residual is scaled down linearly with the deviation: whole where the detector sees healthy tissue, halved at $\tau/2$ and nulled only at $\tau$ or more. The deviation map is computed on the degraded input itself or a second degraded acquisition of the same session as listed above. The acquisition the output is scored against is never read. The modulation moves the brain by at most $0.07$\,dB and lifts the lesion from at or below the input to $+0.25$ (ATLAS-O) and $+0.24$\,dB (BraTS \Tone{}) in the seed mean, $p<10^{-6}$ and $10^{-83}$ paired over cases (\cref{tab:sr}). The control model without pathology randomization, under the same fine-tune, has no detector and is read without modulation.

\subsection{Per contrast inference}
\label[appendix]{sec:suppl_sr_contrast}
\noindent \textbf{Single input.} \Cref{tab:sr_contrast} scores super-resolution on the contrasts of \cref{tab:main} under one rule: the modulation map is the detector's deviation on the network's own degraded input, nothing else of the session is read. Within the brain the modulated output is above the tricubic input on every contrast. Inside the lesion it is above the input wherever the lesion is visible on the input (ATLAS-O \Tone{}, BraTS \FLAIR{} and \Ttwo{}, WMH \FLAIR{}) and below it where the input hides the lesion (BraTS \Tone{}, the WMH \Tone{}s), where the detector cannot modulate what it cannot see; reading the session's \FLAIR{}, degraded like the input, instead lifts BraTS \Tone{} to $+0.24$\,dB (\cref{tab:sr}). Inside the lesion the margin over the control is positive on every contrast except WMH \FLAIR{}, on which both models see the lesion.

\begin{table*}[tbp]
  \centering\small\setlength{\tabcolsep}{3pt}
  \caption{\textbf{Super-resolution per contrast, single-input modulation.} Mean PSNR margin (dB) of detector-modulated \texttt{PathoDraw} over the tricubic input and over the control (no pathology, unmodulated), mean $\pm$ standard deviation over seeds; the modulation map is the detector on the degraded input itself, $\tau=0.4$. BraTS on the acquired-isotropic volumes of each contrast; WMH \FLAIR{} and 2D \Tone{} degraded 3 to 9\,mm, WMH 3D \Tone{} 1 to 3\,mm. Bold: significant gain, paired over cases.}
  \label{tab:sr_contrast}
  \begin{tabular}{@{}ll ccccccc@{}}
    \toprule
    & & ATLAS-O \Tone{} & BraTS \Tone{} & BraTS \FLAIR{} & BraTS \Ttwo{} & WMH \FLAIR{} & WMH \Tone{} & WMH 3D \Tone{} \\
    & & (210) & (690) & (316) & (358) & (60) & (60) & (59) \\
    \midrule
    Over tricubic & Brain  & $\mathbf{+0.96{\pm}0.03}$ & $\mathbf{+0.76{\pm}0.23}$ & $\mathbf{+1.01{\pm}0.10}$ & $\mathbf{+0.96{\pm}0.06}$ & $\mathbf{+0.68{\pm}0.02}$ & $\mathbf{+0.55{\pm}0.03}$ & $\mathbf{+0.89{\pm}0.00}$ \\
                  & Lesion & $\mathbf{+0.25{\pm}0.03}$ & $-0.22{\pm}0.23$ & $\mathbf{+0.52{\pm}0.14}$ & $\mathbf{+0.51{\pm}0.18}$ & $\mathbf{+1.48{\pm}0.22}$ & $-1.11{\pm}0.21$ & $-0.88{\pm}0.03$ \\
    Over control  & Brain  & $\mathbf{+0.32{\pm}0.18}$ & $\mathbf{+0.44{\pm}0.53}$ & $\mathbf{+0.34{\pm}0.28}$ & $-0.31{\pm}0.08$ & $-0.12{\pm}0.06$ & $\mathbf{+0.19{\pm}0.08}$ & $\mathbf{+0.22{\pm}0.01}$ \\
                  & Lesion & $\mathbf{+0.60{\pm}0.28}$ & $\mathbf{+0.28{\pm}0.69}$ & $\mathbf{+0.88{\pm}0.34}$ & $\mathbf{+0.17{\pm}0.33}$ & $-0.02{\pm}0.41$ & $\mathbf{+0.57{\pm}0.19}$ & $\mathbf{+0.66{\pm}0.57}$ \\
    \bottomrule
  \end{tabular}
\end{table*}

\subsection{Deep generative brain prior (\texttt{dmprior})}
\label[appendix]{sec:suppl_sr_dmprior}
The healthy diffusion prior's contrast-specific super-resolution, run on the same degraded pools and scored in the same frame, is below the tricubic input on all but one of the cases it has completed, by $3.7$\,dB within the brain and $5.5$ inside the lesion on ATLAS-O ($170$ of $210$ cases), $4.4$ and $7.2$\,dB on BraTS \Tone{} ($550$ of $690$), $2.4$ and $7.1$\,dB on BraTS \Ttwo{} (all $358$), $1.3$ and $7.0$\,dB on WMH \FLAIR{} ($55$ of $60$; its forward-model fit fails on five Utrecht scans) and $2.7$ and $4.0$\,dB on WMH 3D \Tone{} ($50$ of $59$), means over cases as in \cref{tab:sr}.

\section{Registration}
\label[appendix]{sec:suppl_registration}
The whole-pool comparison of \cref{tab:reg,tab:reg_pretrained} (cases in \cref{fig:reg_mni}) results from \texttt{PathoDraw} and the published pretrained \texttt{SynthMorph} model, an outside reference (\cref{sec:suppl_pretrained}). Additionally, the control backbone without pathology randomization is fine-tuned under \texttt{SynthMorph}'s loss in \cref{tab:reg} and here at equal field size (\cref{tab:reg_lesion}), since a smaller field favors every lesion score. 

\noindent \textbf{Registration augmentation}. Both draws of a pair, two different subjects, are augmented independently before rendering. A residual affine per draw with rotation $\mathcal{U}[-15,15]^{\circ}$, scaling $1+\mathcal{U}[-0.15,0.15]$ and shear $\mathcal{U}[-0.12,0.12]$ per axis, and a smooth random deformation are drawn on a coarse control grid. The registration models double the generator's own setting, a $0.03$--$0.06$ grid at up to 4 voxels standard deviation, to a $0.04$--$0.12$ grid at up to 8 voxels, so that the two draws differ more in shape. The generator's nonlinear field is a Gaussian displacement drawn on the coarse grid and linearly interpolated, a displacement rather than an integrated velocity, so it is not diffeomorphic by construction.

\noindent \textbf{Folds.} The network predicts a stationary velocity field integrated by scaling and squaring, so its field is diffeomorphic up to discretization. A fold is a negative smallest in-brain Jacobian determinant on a case, which the reported field has on $6.3\,\%$ of cases, on fewer than $0.005\,\%$ of the voxels of such a case (median smallest determinant $-0.006$). Evaluation pairs are affinely pre-aligned, and the fine-tune pre-aligns its training pairs with the same routine (300 iterations inside every step). The learned correction is a local field of $1.55$ voxels mean displacement against $2.74$ for the pretrained model at its default weight. The gradient penalty is small on a field of this size, and overlap is maximized with sharp local corrections at structure boundaries, which is where the first-percentile determinant falls ($0.24$ against $0.44$). Pathology in training reduces the folds relative to the control (\cref{tab:reg_lesion}), and $\lambda$ penalizes them directly. \Cref{tab:reg_lambda} sweeps our fine-tune (100k steps, one seed) over \texttt{SynthMorph}'s own weights translated into our loss, each column on the cases \texttt{SynthMorph} completed at that weight. \texttt{SynthMorph} weights the smoothness term by $\lambda_{\mathrm{SM}}$ and the overlap term by $1-\lambda_{\mathrm{SM}}$, with $\lambda_{\mathrm{SM}}=0.5$ as the default, and ours has overlap at weight 1 and smoothness at $\lambda$, so $\lambda=\lambda_{\mathrm{SM}}/(1-\lambda_{\mathrm{SM}})$ and the default is $\lambda=1$. Our field is fold-free from $\lambda=9$ upward (and already at $\lambda=4$, a rung not shown), and \texttt{SynthMorph}'s at every weight. Lesion Dice rises as the field shrinks, while brain alignment falls to the affine level at the stiffest setting. In the cases where our field folds, the lesion-Dice margin over the control model trained without pathology randomization is $+0.008$ and $+0.043$ (seeds 1 and 2), and in the fold-free cases $+0.003$ and $+0.055$, with the Jacobian determinant inside the lesion at $1.02$ in both groups.

\begin{table*}[tbp]
  \centering\small\setlength{\tabcolsep}{6pt}
  \caption{\textbf{Regularization sweep.} Our fine-tune (seed 1) and the published \texttt{SynthMorph} model across the smoothness weight $\lambda_{\mathrm{SM}}$ (ours at $\lambda=\lambda_{\mathrm{SM}}/(1-\lambda_{\mathrm{SM}})$; default $0.5$). Metrics as in \cref{tab:reg_pretrained}; both scored on the cases \texttt{SynthMorph} completed (1{,}877 / 1{,}910 / 1{,}950 / 1{,}217 pairs). Bold: the better of the two at the same weight.}
  \label{tab:reg_lambda}
  \begin{tabular}{@{}ll cccc@{}}
    \toprule
    & $\lambda_{\mathrm{SM}}$ & 0.5 & 0.7 & 0.9 & 0.95 \\
    \midrule
    \texttt{PathoDraw} & Lesion Dice & \textbf{0.827} & \textbf{0.869} & \textbf{0.926} & \textbf{0.885} \\
    & Brain Dice & 0.936 & 0.935 & 0.932 & 0.927 \\
    & $\overline{|u|}$ & 1.55 & 1.19 & 0.80 & 0.97 \\
    & Folds (\%) & 6.3 & 0.7 & \textbf{0.0} & \textbf{0.0} \\
    \addlinespace[2pt]
    \texttt{SynthMorph} & Lesion Dice & 0.703 & 0.749 & 0.852 & 0.873 \\
    & Brain Dice & \textbf{0.949} & \textbf{0.946} & \textbf{0.939} & \textbf{0.934} \\
    & $\overline{|u|}$ & 2.74 & 2.39 & 1.38 & 0.98 \\
    & Folds (\%) & \textbf{0.0} & \textbf{0.0} & \textbf{0.0} & \textbf{0.0} \\
    \bottomrule
  \end{tabular}
\end{table*}

\begin{table*}[tbp]
  \centering\small\setlength{\tabcolsep}{3pt}
  \caption{\textbf{Pathology in registration, against its control.} Ours: the pathology-trained backbone fine-tuned for 100k steps with the pathological rendering on half of the steps. Control: the control model without pathology randomization under the identical fine-tune without pathology randomization, its velocity scaled by $0.97$ to match our mean displacement ($1.53$ against $1.56$ voxels). Lesion Dice as in \cref{tab:reg_pretrained}, with the percentage of cases in which ours is higher; brain Dice, $\det J$ $\wp_{1}$ and folds as there. Bold: the better of the pair; neither where the two are within $0.002$. One seed, 1{,}964 cases.}
  \label{tab:reg_lesion}
  \begin{tabular}{@{}l r cc c ccc cc cc@{}}
    \toprule
    & & \multicolumn{3}{c}{Lesion Dice} & \multicolumn{3}{c}{Brain Dice} & \multicolumn{2}{c}{$\det J$ $\wp_{1}$} & \multicolumn{2}{c}{Folds (\%)} \\
    \cmidrule(lr){3-5}\cmidrule(lr){6-8}\cmidrule(lr){9-10}\cmidrule(lr){11-12}
    & $n$ & Ours & Control & $\Delta$ (\%) & Affine & Ours & Control & Ours & Control & Ours & Control \\
    \midrule
    ATLAS \Tone{}   &  655 & \textbf{0.693} & 0.688 & $+0.005$ (48) & 0.921 & \textbf{0.925} & 0.920 & \textbf{0.263} & 0.189 & \textbf{6.4} & 23.1 \\
    BraTS \Tone{}   & 1{,}249 & 0.917 & \textbf{0.922} & $-0.005$ (26) & 0.937 & 0.940 & 0.939 & \textbf{0.223} & 0.180 & \textbf{5.6} & 35.1 \\
    WMH \FLAIR{}    &   60 & \textbf{0.537} & 0.525 & $+0.012$ (57) & 0.921 & \textbf{0.924} & 0.920 & \textbf{0.248} & 0.183 & \textbf{20.0} & 40.0 \\
    All             & 1{,}964 & 0.831 & 0.832 & $-0.001$ (34) & 0.931 & 0.934 & 0.932 & \textbf{0.236} & 0.183 & \textbf{6.3} & 31.3 \\
    \bottomrule
  \end{tabular}
\end{table*}

\noindent \textbf{Per-structure anatomy.} Pathology randomization does not degrade anatomical accuracy. Carried through each field to \texttt{MNI152} and scored as Dice of the anatomy head's 56-class labels against the template's, 23 large structures (brainstem, and per hemisphere cerebral white matter, cortex, lateral ventricle, cerebellar white matter and cortex, thalamus, caudate, putamen, pallidum, hippocampus, amygdala) score $0.743$ under our field and $0.743$ under the matched control's over 1{,}964 cases (affine $0.671$; paired difference $-0.000$, interval $-0.001$ to $+0.000$), within $0.002$ on every dataset. On the 35-structure set that adds twelve small structures, the means are $0.688$ against $0.691$, within $0.003$ on every dataset at a single seed.

\noindent \textbf{Matched displacement.} Lesion Dice through the warp rewards a smaller field, so a comparison at equal mean displacement can be met by stiffening the pretrained model. Its lesion Dice rises monotonically as its field shrinks (\cref{tab:reg_lambda}). Matching our $1.55$ voxels falls between $\lambda_{\mathrm{SM}}=0.85$ ($1.74$ voxels) and $0.9$ ($1.38$ voxels), and at $0.95$ the warp is sub-voxel. Such a match tests stiffness rather than registration, so we compare the pretrained model at its published default and across the sweep. The control, trained without pathology randomization, is matched to our displacement instead (\cref{tab:reg_lesion}). The stiffer weight is chosen after the fact against the lesion score on the evaluation pool, unavailable a priori. Our field arrives at its size from training alone, at the same published default. At every weight the two were run at, and at equal field size at $\lambda_{\mathrm{SM}}=0.95$, our field keeps more of the lesion without imposed stiffness (\cref{tab:reg_lambda}).

\noindent \textbf{Field conditioning.} At matched displacement (\cref{tab:reg_lesion}) the two fields keep the lesion and align the brain alike. Pathology randomization improves the conditioning of the field: the first-percentile determinant is $0.24$ against $0.18$ and the field folds on $6.3\,\%$ of cases against $31.3\,\%$.

\subsection{Pretrained SynthMorph}
\label[appendix]{sec:suppl_pretrained}
The published \texttt{SynthMorph} model is pretrained (\cref{tab:reg_pretrained}) and inferred on the test datasets at its default regularization. Its deformable model is a hypernetwork whose regularization weight is a test-time parameter ($\lambda_{\mathrm{SM}}\in(0,1)$ in its notation, the weight of the smoothness term against $1-\lambda_{\mathrm{SM}}$ on the label overlap \cite{synthmorph_joint}; default 0.5). The \texttt{HyperMorph} extension \cite{hypermorph} features amortized selection that could choose $\lambda_{\mathrm{SM}}$ automatically, but only against a lesion criterion on annotated held-out scans, which a zero-shot setting lacks. The \texttt{SynthMorph} field at the default weight is larger than ours ($2.74$ against $1.55$ voxels). At $\lambda_{\mathrm{SM}}=0.9$ it halves its field ($1.38$ voxels) and reaches lesion Dice $0.85$, above our $0.83$ at the default weight but below our $0.93$ at the same weight (\cref{tab:reg_lambda}), and its field is better conditioned at the default weight (\cref{tab:reg_pretrained}) and fold-free at every weight (\cref{tab:reg_lambda}).

\begin{table*}[tbp]
  \caption{\textbf{Pretrained \texttt{SynthMorph}} \cite{synthmorph,synthmorph_joint} (\texttt{mri\_synthmorph}, FreeSurfer 8, default $\lambda_{\mathrm{SM}}=0.5$) against our field on every ATLAS, BraTS and WMH case with a lesion mask (1{,}877 pairs), registered to \texttt{MNI152} from the same affine. Lesion Dice: overlap of the lesion carried through the field with its affine-only position, mean, with paired difference and Wilcoxon $p$. Brain Dice: overlap of the warped brain mask with the template's, mean; Affine is the same before the deformable step. $\overline{|u|}$: mean displacement in the brain (voxels). $\det J$ $\wp_{1}$: first percentile of the Jacobian determinant in the brain, median over cases. Folds: cases with a negative determinant. Bold: the better of the two.}
  \label{tab:reg_pretrained}
  \centering\small
  \setlength{\tabcolsep}{4pt}
  \begin{tabular}{@{}lr ccc r ccc cc cc cc@{}}
    \toprule
    & & \multicolumn{3}{c}{Lesion Dice} & & \multicolumn{3}{c}{Brain Dice} & \multicolumn{2}{c}{$\overline{|u|}$} & \multicolumn{2}{c}{$\det J$ $\wp_{1}$} & \multicolumn{2}{c}{Folds (\%)} \\
    \cmidrule(lr){3-5}\cmidrule(lr){7-9}\cmidrule(lr){10-11}\cmidrule(lr){12-13}\cmidrule(lr){14-15}
    & $n$ & Ours & \texttt{SM} & $\Delta$ & $p$ & Affine & Ours & \texttt{SM} & Ours & \texttt{SM} & Ours & \texttt{SM} & Ours & \texttt{SM} \\  %
    \midrule
    ATLAS \Tone{}  &  639 & \textbf{0.693} & 0.588 & $+0.105$ & $10^{-62}$  & 0.921 & 0.925 & 0.925 & 1.46 & 2.70 & 0.261 & \textbf{0.420} & 6.4 & \textbf{0.0} \\
    BraTS          & 1{,}179 & \textbf{0.915} & 0.782 & $+0.133$ & $10^{-192}$ & 0.940 & 0.943 & \textbf{0.963} & 1.61 & 2.77 & 0.225 & \textbf{0.447} & 5.5 & \textbf{0.0} \\
    WMH \FLAIR{}   &   59 & \textbf{0.537} & 0.384 & $+0.153$ & $10^{-11}$  & 0.922 & \textbf{0.924} & 0.916 & 1.48 & 2.62 & 0.247 & \textbf{0.432} & 20.3 & \textbf{0.0} \\
    All            & 1{,}877 & \textbf{0.827} & 0.703 & $+0.124$ & $10^{-264}$ & 0.933 & 0.936 & \textbf{0.949} & 1.55 & 2.74 & 0.238 & \textbf{0.437} & 6.3 & \textbf{0.0} \\
    \bottomrule
  \end{tabular}
\end{table*}

\section{Training details}
\label[appendix]{sec:suppl_training}

\noindent \textbf{Detection.} The reported model is the recipe of \cref{sec:training} trained from random weights to 400k steps (two seeds), then continued for 100k steps with the cosine decay and the exponential moving average (EMA) to 500k.
The component ablation of \cref{tab:ablation} branches at 300k from the seed-matched routine runs and trains each branch to 400k, so its rows are paired forks of one ancestor and are read from raw weights at 400k. The data-regime ablation of \cref{tab:regime} is read at the 500k budget from the EMA weights, as \cref{tab:main}. The control models differ from the reported model only by the absence of pathology randomization.

\noindent \textbf{Fine-tuning.} Super-resolution and registration fine-tune at constant learning rate $10^{-4}$. Half of the super-resolution draws are skull-stripped. For registration the two draws of a pair receive independent residual affines and the generator's smooth deformation at twice its default amplitude and are affinely pre-aligned before the network, as at test time (\cref{sec:suppl_registration}). 

\section{Neurodegeneration}
\label[appendix]{sec:suppl_adni}
ADNI3~\cite{adni3} (331 scans per contrast, \Tone{} and \FLAIR{}; 132 cognitively normal, 199 impaired) is a neurodegenerative cohort without focal lesions. Dice between the anatomy head and the control's, mean over classes, median scan, mean over seeds, is 0.857 / 0.858 (\Tone{}) and 0.826 / 0.829 (\FLAIR{}) for cognitively normal / impaired subjects.

\section{Benchmark context}
\label[appendix]{sec:suppl_benchmark}
\noindent The published \texttt{FAE} and \texttt{DISYRE} values were scored on the benchmark's case lists, the BraTS 2020 training subjects and 54 of the 60 WMH subjects, and \cref{tab:main} scores \texttt{PathoDraw} on those lists through the organizers' name mapping (329 of the 332 BraTS subjects are in the 2023 pool with a histogram; 54 of 54 WMH), threshold re-selected on the subset: $0.537$ [$0.516$, $0.558$] on BraTS \Tone{} (every case of BraTS 2023: $0.539$), $0.622$ [$0.600$, $0.642$] on BraTS \FLAIR{} ($0.654$), $0.657$ [$0.635$, $0.678$] on BraTS \Ttwo{} ($0.653$), $0.259$ [$0.206$, $0.314$] on WMH \FLAIR{} (all 60: $0.268$) and $0.229$ [$0.178$, $0.281$] on WMH \Tone{} ($0.239$); ATLAS-O is the benchmark's set already. The comparison is unchanged: above \texttt{FAE} on BraTS \Tone{} and above \texttt{DISYRE} on WMH \Tone{} outside the interval, and level with \texttt{DISYRE} on BraTS \Ttwo{} inside it.
\texttt{DISYRE}~\cite{disyre,disyrev2} has reported Dice scores up to 0.45 on
ATLAS, 0.51 on BraTS \Tone{} and 0.73 on BraTS \Ttwo{}. Those compute thresholds by a different method than the independent benchmark. The published evaluation pools every voxel of the test set before searching the threshold, so its ceiling Dice is dominated by the largest lesions, computed on 2D slices over all 655 ATLAS scans with no scanner split and with that method's own preprocessing. 
The benchmark~\cite{uadbench} instead averages a per-individual Dice at one global threshold, which is what every row of \cref{tab:main} reports. For reference only, the same pooled ceiling computed from our per-case histograms (4\,mm median filter) is 0.53 on all ATLAS cases (seeds 0.55 and 0.51), 0.59 on BraTS \Tone{} (0.59 and 0.59) and 0.72 on BraTS \Ttwo{} (0.73 and 0.71), against a per-individual 0.34, 0.54 and 0.65.

\section{Confidence intervals}
\label[appendix]{sec:suppl_pooled}
\noindent 95\,\% bootstrapped confidence intervals of the \texttt{PathoDraw} column of \cref{tab:main}
(20{,}000 resamples of the cases, seed-averaged per-case Dice at the global threshold) are, on the benchmark's case lists, ATLAS-O 0.34--0.40, BraTS \Tone{} 0.52--0.56,
BraTS \FLAIR{} 0.60--0.64, BraTS \Ttwo{} 0.64--0.68, WMH \FLAIR{} 0.21--0.31 and WMH \Tone{} 0.18--0.28. The published methods are single
point estimates, so a value is bolded where the published value lies below our interval (BraTS \Tone{} against \texttt{FAE}, WMH \Tone{} against
\texttt{DISYRE}) or inside it (ATLAS-O and BraTS \Ttwo{} against \texttt{DISYRE}). No published value lies above our intervals.

\begin{table*}[tbp]
  \centering\scriptsize\setlength{\tabcolsep}{2.5pt}
  \caption{\textbf{Intervals and $p$-values behind the main tables.} Against a published value (one number, other cases): 95\,\% bootstrap interval of the seed-averaged per-case Dice (20{,}000 resamples) and the one-sided bootstrap $p$ that \texttt{PathoDraw}'s mean is at or below the published value. Against a model scored on the same cases: paired per-case difference, Wilcoxon signed-rank $p$. \texttt{PathoDraw} on the benchmark's case lists for the detection rows. $^\dagger$Super-resolution rows list ATLAS-O / BraTS \Tone{} / BraTS \Ttwo{} / WMH \FLAIR{}; a negative margin is tested in its own direction. Median-scan match: the cutoff at which the median healthy scan, rather than the mean, flags the published rate.}
  \label{tab:pvalues}
  \begin{tabular}{@{}l l r l l r@{}}
    \toprule
    Table & Comparison & $n$ & \texttt{PathoDraw} [95\,\% CI] & Other & $p$ \\
    \midrule
    \cref{tab:main} & ATLAS-O \Tone{} vs \texttt{DISYRE} & 210 & 0.369 [0.337, 0.400] & 0.386 & 0.86 \\
    & BraTS \Tone{} vs \texttt{FAE} & 329 & 0.537 [0.516, 0.558] & 0.437 & $<10^{-4}$ \\
    & BraTS \Ttwo{} vs \texttt{DISYRE} & 329 & 0.657 [0.635, 0.678] & 0.646 & 0.17 \\
    & WMH \Tone{} vs \texttt{DISYRE} & 54 & 0.229 [0.178, 0.281] & 0.130 & $10^{-4}$ \\
    & Control (no pathology), paired, six rows & 54--329 & As in \cref{tab:main} & As in \cref{tab:main} & $<10^{-7}$ on five rows; WMH \FLAIR{} 0.45 \\
    & \texttt{dmprior} (no pathology), paired, six rows & 54--329 & As in \cref{tab:main} & As in \cref{tab:main} & $<10^{-9}$ on every row \\
    \midrule
    \cref{tab:matched_fpr} & ATLAS-O at \texttt{FAE}'s specificity & 210 & 0.364 [0.332, 0.397] & 0.166 & $<10^{-4}$ \\
    & BraTS at \texttt{FAE}'s & 329 & 0.423 [0.402, 0.444] & 0.437 & 0.91 \\
    & WMH at \texttt{FAE}'s & 54 & 0.229 [0.178, 0.281] & 0.016 & $<10^{-4}$ \\
    & ATLAS-O at \texttt{DISYRE}'s & 210 & 0.269 [0.244, 0.296] & 0.384 & 1.00 \\
    & BraTS at \texttt{DISYRE}'s & 329 & 0.160 [0.145, 0.175] & 0.264 & 1.00 \\
    & WMH at \texttt{DISYRE}'s & 54 & 0.102 [0.073, 0.134] & 0.109 & 0.67 \\
    & Median-scan match: ATLAS-O at \texttt{FAE}'s & 210 & 0.363 [0.330, 0.396] & 0.166 & $<10^{-4}$ \\
    & BraTS at \texttt{FAE}'s & 329 & 0.427 [0.407, 0.448] & 0.437 & 0.82 \\
    & WMH at \texttt{FAE}'s & 54 & 0.229 [0.178, 0.281] & 0.016 & $<10^{-4}$ \\
    & ATLAS-O at \texttt{DISYRE}'s & 210 & 0.316 [0.288, 0.344] & 0.384 & 1.00 \\
    & BraTS at \texttt{DISYRE}'s & 329 & 0.214 [0.198, 0.231] & 0.264 & 1.00 \\
    & WMH at \texttt{DISYRE}'s & 54 & 0.140 [0.103, 0.179] & 0.109 & 0.053 \\
    \midrule
    \cref{tab:sr} & Against tricubic, brain$^\dagger$ & 210/690/358/60 & $+0.96$/$+0.79$/$+1.09$/$+0.68$ & Paired & $10^{-33}$/$10^{-107}$/$10^{-60}$/$10^{-11}$ \\
    & Against tricubic, lesion & 210/690/358/60 & $+0.25$/$+0.24$/$+0.44$/$+1.48$ & Paired & $10^{-6}$/$10^{-83}$/$10^{-59}$/$10^{-9}$ \\
    & Against control, brain & 210/690/358/60 & $+0.32$/$+0.47$/$-0.18$/$-0.12$ & Paired & $10^{-36}$/$10^{-112}$/$10^{-34}$/$10^{-10}$ \\
    & Against control, lesion & 210/690/358/60 & $+0.60$/$+0.74$/$+0.10$/$-0.02$ & Paired & $10^{-29}$/$10^{-102}$/0.9/0.8 \\
    & Against \texttt{dmprior} (no pathology), brain & 170/550/358/55 & $+4.7$/$+5.2$/$+3.5$/$+2.0$ & Paired & $10^{-29}$/$10^{-92}$/$10^{-60}$/$10^{-10}$ \\
    & Against \texttt{dmprior} (no pathology), lesion & 170/550/358/55 & $+5.8$/$+7.5$/$+7.6$/$+8.6$ & Paired & $10^{-29}$/$10^{-92}$/$10^{-60}$/$10^{-10}$ \\
    \midrule
    \cref{tab:reg} & Lesion Dice vs control & 210 / 1{,}249 / 60 & $+0.046$ / $+0.014$ / $+0.063$ & Paired & $10^{-30}$ / $10^{-179}$ / $10^{-10}$ \\
    & Lesion Dice vs \texttt{SynthMorph} & 209 / 1{,}179 / 59 & $+0.112$ / $+0.140$ / $+0.180$ & Paired & $10^{-25}$ / $10^{-192}$ / $10^{-11}$ \\
    \cref{tab:reg_lesion} & Lesion Dice vs matched control & 655 / 1{,}249 / 60 & $+0.005$ / $-0.005$ / $+0.012$ & Paired & 0.42 / $10^{-71}$ / 0.038 \\
    \bottomrule
  \end{tabular}
\end{table*}

\section{Healthy cohorts}
\label[appendix]{sec:suppl_healthy}

\label[appendix]{sec:healthy}The healthy cohort is HCP-Aging~\cite{hcpaging}, 2{,}092 scans per contrast, none of them in training. Dice between the anatomy head and the control's, mean over classes, median scan, mean over seeds, is 0.906 / 0.898 (\Tone{} / \Ttwo{}).

\subsection{Matched specificity}
\label[appendix]{sec:suppl_fpr}
\cref{tab:matched_fpr} reads \texttt{PathoDraw} at each published model's healthy flag rate, with the threshold set on HCP-Aging and Dice on the benchmark's case lists. This invokes a comparison with \texttt{FAE} and \texttt{DISYRE} at matched specificity.
\begin{table*}[tbp]
  \caption{\textbf{Dice at matched healthy specificity.} \texttt{PathoDraw} read at each published model's healthy false-positive rate (cutoff set on HCP-Aging), on the benchmark's case lists (ATLAS-O 210, BraTS 329, WMH 54), with 95\,\% bootstrap intervals. Published: \texttt{FAE} and \texttt{DISYRE} Dice at their own healthy thresholds (median-filtered), not the ceiling Dice of \cref{tab:main}; \Tone{} rows. Level from: the lowest healthy flag rate at which the published value lies inside or below \texttt{PathoDraw}'s interval. Bold: inside or above the published interval; $p$-values in \cref{tab:pvalues}.}
  \label{tab:matched_fpr}
  \centering\footnotesize\setlength{\tabcolsep}{5pt}
  \begin{tabular}{l c | cc | cc}
    \toprule
    & Published & \multicolumn{2}{c|}{At \texttt{FAE}'s rate ($0.82\,\%$)} & \multicolumn{2}{c}{At \texttt{DISYRE}'s rate ($0.12\,\%$)} \\
    & [min, max] & \texttt{PathoDraw} & Level from & \texttt{PathoDraw} & Level from \\
    \midrule
    ATLAS-O     & [0.17, 0.38] & \textbf{0.36} [0.33, 0.40] & $0.04\,\%$ & \textbf{0.27} [0.24, 0.30] & $0.29\,\%$ \\
    BraTS       & [0.26, 0.44] & \textbf{0.42} [0.40, 0.44] & $0.78\,\%$ & 0.16 [0.15, 0.18] & $0.21\,\%$ \\
    WMH         & [0.02, 0.11] & \textbf{0.23} [0.18, 0.28] & $0.03\,\%$ & \textbf{0.10} [0.07, 0.13] & $0.10\,\%$ \\
    \midrule
    Mean        & [0.15, 0.31] & 0.34 & & 0.18 & \\
    \bottomrule
  \end{tabular}
\end{table*}

\subsection{False positives}
\label[appendix]{sec:failure}

False positives and negatives are described in \cref{sec:suppl_failure}.

\section{Failure analysis}
\label[appendix]{sec:suppl_failure}

False positives on every dataset are CSF extremes plus mildly deviant white matter, dominant on ATLAS. Dark CSF voxels are the majority of false-positive volume on WMH \FLAIR{} and PRES scans. We expect this is because synthetic CSF is modeled as one mean per label while a real ventricle is darker at its center than at its walls. Misses are intermediate-intensity rim and edema that the anatomy head labels cortex, so the robust tissue-class statistic scores them as normal primarily on BraTS \Ttwo{} and \FLAIR{}.

\section{Manifold enclosure}
\label[appendix]{sec:suppl_manifold}

\begin{figure*}[t]
  \centering
  \includegraphics[width=\linewidth,height=0.42\textheight,keepaspectratio]{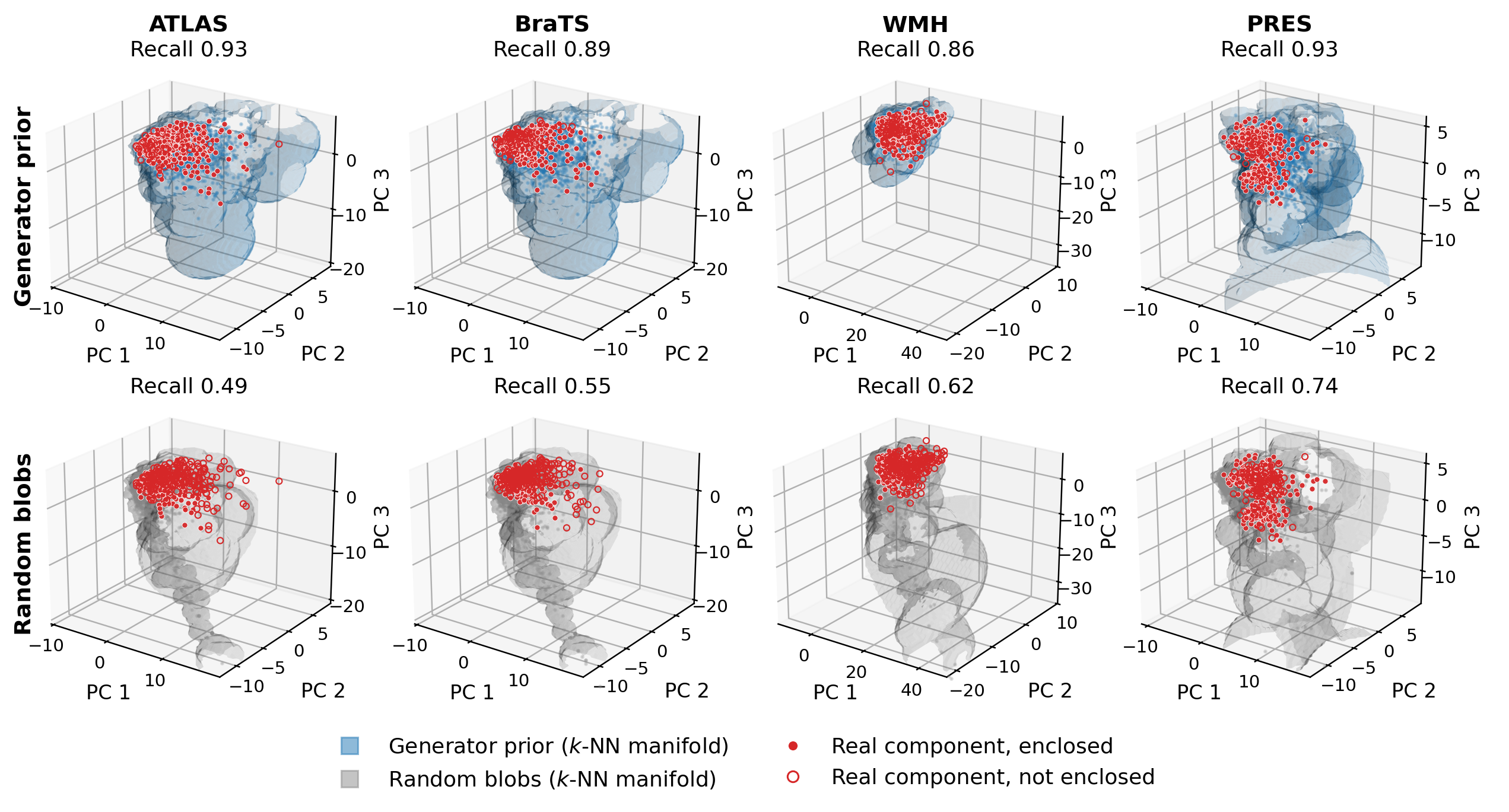}
  \caption{\textbf{Manifold enclosure in the Zernike shape space}: the generator's $k$-NN manifold ($k=5$ \cite{kynkaanniemi}, translucent) and random blobs (second row), projected on the first three principal components of the 1\,mm draws, against each dataset's real lesion components on its matched voxel grid (filled when the full 36-D test encloses them, hollow when not; titles give the recall), ATLAS and BraTS sharing the 1\,mm generator set by construction.}
  \label{fig:manifold}
\end{figure*}

\begin{figure*}[t]
  \centering
  \includegraphics[width=\linewidth,height=0.4\textheight,keepaspectratio]{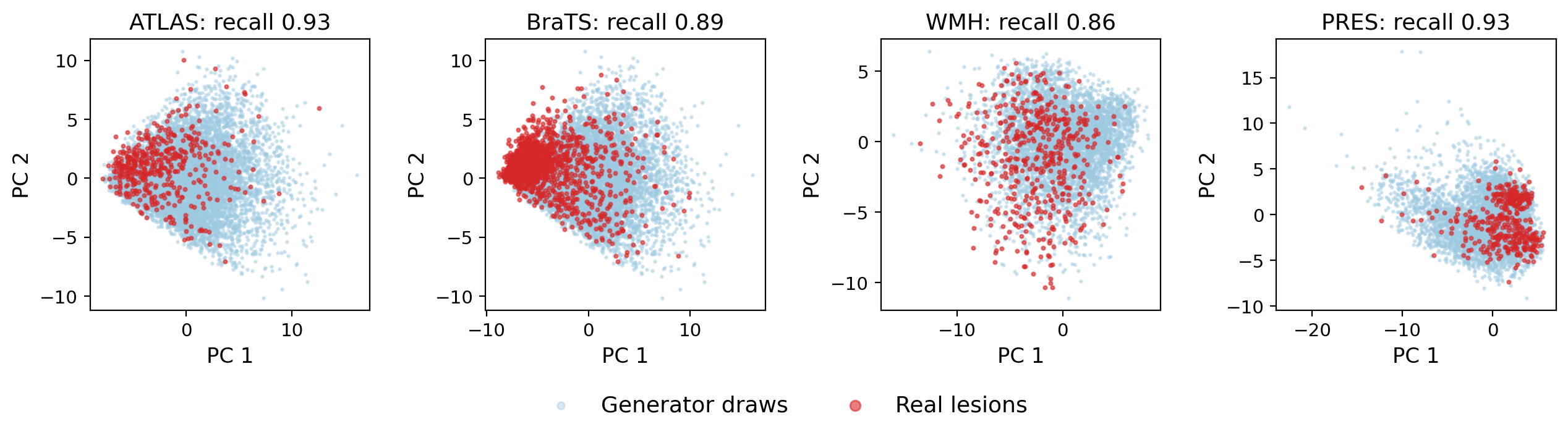}
  \caption{Descriptor space (first two principal components of the synthetic set) with the real lesion components of each dataset, on matched voxel grids; titles give the $k=5$ recall.}
  \label{fig:manifold2d}
\end{figure*}

Independent descriptors include 3D Zernike moments $\hat{\Omega}_{nlm}$ of each 26-connected component scaled into the unit ball, rotation invariants $F_{nl}=\|\hat{\Omega}_{nl}\|_2$ for $n\le10$ (36 values), and volume reported separately. For a component $\mathcal{C}$ with voxel positions $\mathbf{p}_v$, centered at its center of mass and scaled so that $\max_v\|\mathbf{p}_v\|=1$, the canonical 3D Zernike functions~\cite{novotni2004} are
  \begin{equation}
  \begin{aligned}
    Z_{nlm}(\mathbf{p})&=R_{nl}(\rho)\,Y_{lm}(\hat{\mathbf{p}}),\\
    R_{nl}(\rho)&=\textstyle\sum_{\nu=0}^{\kappa}q_{\kappa l\nu}\,\rho^{2\nu+l},\\
    q_{\kappa l\nu}&=\frac{(-1)^{\kappa+\nu}}{2^{2\kappa}}\sqrt{\frac{2l+4\kappa+3}{3}}\,\binom{2\kappa}{\kappa}\binom{\kappa}{\nu}\frac{\binom{2(\kappa+l+\nu)+1}{2\kappa}}{\binom{\kappa+l+\nu}{\kappa}},\\
    \hat{\Omega}_{nlm}&=\tfrac{3}{4\pi}\textstyle\sum_{v\in\mathcal{C}}w_v\,\overline{Z_{nlm}(\mathbf{p}_v)},\\
    F_{nl}&=\big(\textstyle\sum_{m=-l}^{l}|\hat{\Omega}_{nlm}|^{2}\big)^{1/2},
  \end{aligned}
  \label{eq:zernike}
  \end{equation}
with $n\le10$ the order, $l\in\{0,\dots,n\}$ the degree with $n-l=2\kappa$ even, $m\in\{-l,\dots,l\}$ the azimuthal index, $\nu$ the summation index of the radial polynomial $R_{nl}$, $\rho=\|\mathbf{p}\|$ the radius and $\hat{\mathbf{p}}=\mathbf{p}/\rho$ the direction of a position, $Y_{lm}$ the spherical harmonics, $\overline{\,\cdot\,}$ the complex conjugate, $w_v$ the volume of voxel $v$ normalized to sum to one over $\mathcal{C}$, and $\hat{\Omega}_{nl}=(\hat{\Omega}_{nlm})_{m=-l}^{l}$ the moments of one $(n,l)$ pair. The 36 pairs $(n,l)$ with $n\le10$ give the 36 invariants.
 The descriptors were z-scored on the synthetic set. Components of at least $0.03$\,mL (30 voxels at 1\,mm) enter the analysis, and any component larger than 6{,}000 voxels ($6$\,mL at 1\,mm) is subsampled to 6{,}000 voxels. To estimate manifold enclosure, $k$-NN precision and recall \cite{kynkaanniemi} with $k=5$ were used. Synthetic masks were block-resampled to the real grid before description. The control, fixed a priori, consisted of 2{,}500 random blobs (unions of 1--4 random ellipsoids (axis ratios $\mathcal{U}[1,3]$, random rotation, random offsets, largest connected component kept), volumes log-uniform in $[0.03, 200]$\,mL) described on each dataset's grid like the draws (\cref{fig:manifold,fig:manifold2d}). Single ellipsoids were also tested and enclose nothing (recall $<0.01$), as their descriptors lie on a two-parameter curve whose nearest-neighbor radii are too small to contain any real shape, so they are omitted.

\subsection{Coverage measurement}
\label[appendix]{sec:coverage}
Every lesion component of the evaluation masks and of 1{,}000 generator draws is described by its 3D Zernike invariants~\cite{novotni2004}, and a real component counts as enclosed when it lies within the $k$-nearest-neighbor ball of a generator component~\cite{kynkaanniemi} (\cref{tab:manifold}): recall is $0.86$--$0.93$ on all four datasets with precision $0.87$--$0.95$.

\subsection{Enclosure interpretation}
\label[appendix]{sec:suppl_enclosure}

A control shape distribution, specified without reference to the generator or to any dataset (volumes log-uniform in 0.03--200\,mL; construction in \cref{tab:manifold}), bounds the reading: random blobs, unions of one to four random ellipsoids (axis ratios in $[1,3]$), enclose 0.49--0.74 of the real components with precision 0.57--0.73, against 0.86--0.93 and 0.87--0.95 for the generator. Shape enclosure is therefore necessary but not sufficient. A shape-only null covers half of the real shapes, the anatomy-conditioned modes account for the rest, and enclosure does not order the Dice results (ATLAS has the highest enclosure and a mid-range Dice). The residual loss is in appearance, which \cref{sec:suppl_failure} locates in the healthy rendering.

\begin{table}[htbp]
  \caption{Manifold enclosure of real lesion components in the 3D Zernike descriptor space ($k=5$), one generator reference set per voxel grid (ATLAS and BraTS are both 1\,mm and therefore share it): recall = real components inside the generator's $k$-NN manifold, precision = the converse; synthetic masks resampled to each dataset's voxel grid (ATLAS, BraTS 1\,mm; WMH $1\times1\times3$\,mm; PRES $0.6\times0.6\times6$\,mm), components $\ge0.03$\,mL. Control: 2{,}500 random blobs (unions of 1--4 random ellipsoids, volumes log-uniform in 0.03--200\,mL; no parameter taken from the generator or the data).}
  \label{tab:manifold}
  \centering\small
  \setlength{\tabcolsep}{4pt}
  \begin{tabular}{@{}lccccc@{}}
    \toprule
    & & \multicolumn{2}{c}{Generator} & \multicolumn{2}{c}{Random blobs} \\
    \cmidrule(lr){3-4}\cmidrule(lr){5-6}
    Dataset & $n$ & Recall & Precision & Recall & Precision \\
    \midrule
    ATLAS & 448 & \textbf{0.93} & \textbf{0.88} & 0.49 & 0.57 \\
    BraTS & 1{,}833 & \textbf{0.89} & \textbf{0.91} & 0.55 & 0.65 \\
    WMH & 475 & \textbf{0.86} & \textbf{0.95} & 0.62 & 0.70 \\
    PRES & 387 & \textbf{0.93} & \textbf{0.87} & 0.74 & 0.73 \\
    \bottomrule
  \end{tabular}
\end{table}

\subsection{Effect of lesion scale}
\label[appendix]{sec:suppl_lesion_scale}
The compared published methods trained only on healthy priors do not corrupt pathology uniformly. We speculate that a prior fit to healthy anatomy represents any structure on the scale of that anatomy, a stroke cavity like a ventricle, whereas a punctate lesion has no healthy counterpart at its scale and is regressed to the surrounding tissue.

\end{document}